\documentclass[10pt,twocolumn,letterpaper]{article}

\usepackage[pagenumbers]{cvpr} 

\usepackage[dvipsnames]{xcolor}

\usepackage[utf8]{inputenc}
\usepackage{amsfonts,amsmath}
\usepackage{color}
\usepackage{tabularx}
\usepackage[most]{tcolorbox}
\usepackage{float}
\usepackage{afterpage}
\usepackage{cuted}
\usepackage{booktabs} 
\usepackage{xspace}
\usepackage{enumitem}
\usepackage{graphicx}
\usepackage[ruled]{algorithm2e} 
\usepackage{algorithmicx}
\algrenewcommand\algorithmicindent{0.5em}
\usepackage{algpseudocode}
\usepackage{nicefrac}
\setlist[itemize]{leftmargin=*,noitemsep}
\usepackage{makecell}
\usepackage{relsize}
\usepackage{stfloats}
\usepackage[labelfont={bf,it},textfont=it]{caption}
\usepackage{annotate-equations} 
\usepackage{subcaption}
\newsavebox{\measurebox}
\usepackage{mathtools}
\usepackage{xspace}
\usepackage{esvect}
\usepackage{accents}
\usepackage{colortbl}
\usepackage{array}
\usepackage{hhline}
\usepackage{nicematrix}
\usepackage{siunitx}
\usepackage{multirow}
\usepackage{placeins}
\usepackage{adjustbox}
\usepackage{rotating}
\usepackage{balance}
\usepackage{pifont}%
\usepackage[ruled]{algorithm2e} 
\usepackage{calc}
\usepackage[pagebackref,breaklinks,colorlinks,citecolor=cvprblue]{hyperref}

\definecolor{lightheadergrey}{RGB}{240,240,240}
\definecolor{lightblue}{RGB}{235,244,250}
\definecolor{lighterblue}{RGB}{242,248,252}
\definecolor{darkblue}{RGB}{188, 203, 220}
\definecolor{mylightred}{RGB}{222, 146, 146}
\colorlet{mylightredtrans}{mylightred!40}
\definecolor{algosection}{RGB}{240,240,240}
\definecolor{algosectiondarker}{RGB}{220,220,220}
\definecolor{commentcolor}{RGB}{0,128,0}
\definecolor{sectioncolor}{RGB}{255,0,0}
\definecolor{cvprblue}{rgb}{0.21,0.49,0.74}
\definecolor{ForestGreen}{rgb}{0.13, 0.55, 0.13}

\SetAlFnt{\small}
\SetAlCapFnt{\small}
\SetAlCapNameFnt{\small}
\SetAlCapHSkip{0pt}

\makeatletter
\DeclareRobustCommand\onedot{\futurelet\@let@token\@onedot}
\def\@onedot{\ifx\@let@token.\else.\null\fi\xspace}
\def\eg{{e.g}\onedot} 
\def\ie{{i.e}\onedot}

\def\etal{\emph{et al}\onedot}
\makeatother

\makeatletter
\newcommand\GPKern[1]{%
    \@ifnextchar\bgroup{\GPKern@double{#1}}{\GPKern@single{#1}}%
}
\newcommand\GPKern@single[1]{%
    {\ensuremath{\mathbf{K}_{ \mathbf{#1} }^{} } }%
}
\newcommand\GPKern@double[2]{%
    {\ensuremath{\mathbf{K}_{ \mathbf{#1} }^{\ #2} } }%
}
\makeatother

\MakeRobust{\vv}

\newcommand{\FlowField}{
     \ensuremath{ \vv{\mathbf{f}} }
}

\SetAlFnt{\small}
\SetAlCapFnt{\small}
\SetAlCapNameFnt{\small}
\SetAlCapHSkip{0pt}

\MakeRobust{\vv}

\newcommand{\cellline}{\raisebox{0.75ex}{\rule{0.66em}{1.0pt}}}

\newcommand{\cmark}{{\color{ForestGreen} \ding{51}}}%
\newcommand{\xmark}{{\color{red} \ding{55}}}%
\newcommand{\cmarkna}{{\color{gray!50} \cellline}}%

\newcommand{\rom}[1]{\uppercase\expandafter{\romannumeral #1\relax}}

\DeclareMathAlphabet\mathbfcal{OMS}{cmsy}{b}{n}

\colorlet{AlgCaptionColor}{darkblue}

\makeatletter
\renewcommand{\algocf@makecaption@ruled}[2]{%
  \global\sbox\algocf@capbox{%
    \colorbox{AlgCaptionColor}{%
      \makebox[0.97\columnwidth][l]{%
        \hskip\AlCapHSkip%
        \parbox[t]{\dimexpr\columnwidth-\AlCapHSkip}{\algocf@captiontext{\strut#1}{\strut#2\strut}}%
      }%
    }%
  }%
}%
\makeatother

\def\paperID{205} 
\def\confName{3DV\xspace}
\def\confYear{2025\xspace}

\title{ARC-Flow : Articulated, Resolution-Agnostic, Correspondence-Free Matching and Interpolation of 3D Shapes Under Flow Fields \vspace{-20pt}} 
\author{Adam Hartshorne \hspace{1.0em} Allen Paul \hspace{1.0em} Tony Shardlow \hspace{1.0em} Neill D.\,F. Campbell \\
University of Bath \\
\small{\tt ath35, ap2746, tjs42, nc537 @bath.ac.uk}
}

\renewcommand{\paragraph}[1]{\noindent \textbf{#1}:~}
\setlist{nosep}

\begin{document}

\maketitle

\setkeys{Gin}{keepaspectratio}

\begin{strip}
\begin{minipage}{\textwidth}\centering
\vspace{-40pt}
\includegraphics[width=0.96\textwidth]{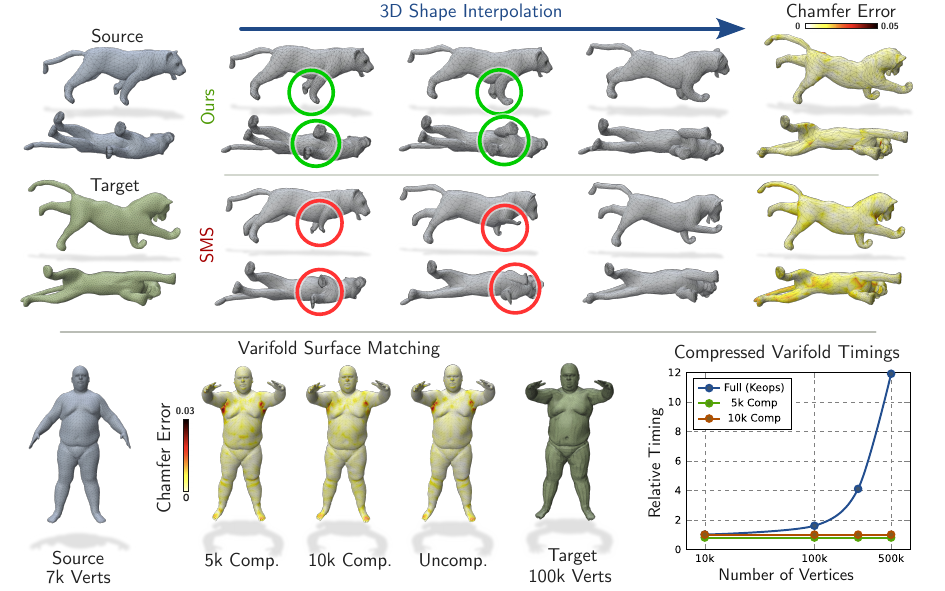}
\vspace{-5pt}
\captionof{figure}{\textbf{Top: 3D shape interpolation.} Our method obtains a more reliable interpolation than existing state-of-the-art  approaches (\eg~Spectral Meets Spatial~\cite{cao2024spectral}) even under substantial non-isometric deformation.
Samples taken at $t = {0.25, 0.5, 0.75}$ and $1.0=:T$ along the deformation path.  
\textbf{Bottom: Scaling to high-resolution meshes.} Using Varifold compression we obtain dramatic computational savings, while maintaining similar perceptual quality, allowing us to scale to high resolution meshes.
}
\vspace{-7.0pt}
\label{fig:teaser}
\end{minipage}
\end{strip}

\flushbottom
\begin{abstract}
\vspace{-7.5pt}
{\hyphenpenalty=10000 
This work presents a unified framework for the unsupervised prediction of physically plausible interpolations between two 3D articulated shapes and the automatic estimation of dense correspondence between them.
Interpolation is modelled as a diffeomorphic transformation using a smooth, time-varying flow field governed by Neural Ordinary Differential Equations (ODEs). \looseness=-1 This ensures topological consistency and non-intersecting trajectories while accommodating hard constraints, such as volume preservation, and soft constraints, \eg physical priors. 

Correspondence is recovered using an efficient Varifold formulation, that is effective on high-fidelity surfaces with differing parameterisations. 
By providing a simple skeleton for the source shape only, we impose physically motivated constraints on the deformation field and resolve symmetric ambiguities. This is achieved without relying on skinning weights or any prior knowledge of the skeleton's target pose configuration.
Qualitative and quantitative results demonstrate competitive or superior performance over existing state-of-the-art approaches in both shape correspondence and interpolation tasks across standard datasets.
}

\end{abstract}
   
\vspace{-7.05pt}
{\small \paragraph{Acknowledgements}} {\small \hspace{-0.5em}  We are grateful for support from the EPSRC SAMBa Centre for Doctoral Training in Statistical Applied Mathematics (EP/L015684/1), the EPSRC CAMERA Research Centre (EP/M023281/1 and EP/T022523/1), UKRI Strength in Places Fund MyWorld Project
(SIPF00006/1) and the Royal Society.}
\newpage
\section{Introduction}
\label{sec:introduction}

\looseness=-3 Recovering maps between 3D shapes is a fundamental problem in computer vision and graphics since it facilitates shape analysis and generation. In particular, modelling realistic non-rigid deformations of 3D shapes is a recognisable problem at the heart of numerous tasks, including animation, shape comparison and style transfer. Although shape matching and interpolation are closely related, they are often treated separately or sequentially, \eg in learning statistical shape models.

Aligning a common template to a set of shapes, such as 3D human point clouds in different poses, is also a long-standing problem crucial for geometric learning. Despite extensive research, establishing accurate correspondences between non-rigidly deformed shapes remains challenging in real-world scenarios due to non-isometric deformations, variations in discretisation, and symmetries, \eg those in bipeds and quadrupeds. Furthermore, there is often a desire to also transfer a skeletal structure from a template to different sample poses \cite{musoni2021functional} for applications such as rigging.

While articulated objects undergo shape changes due to articulation, these changes are not arbitrary. Interpolation involves finding a continuous deformation path from a source shape to a target shape, ensuring that the trajectory represents a realistic articulation of the underlying physical object. Moreover, the shape must remain topologically valid and self-intersection-free throughout the trajectory.

\paragraph{Contributions} This work encompasses the following:
\begin{itemize}
\item \looseness=-2 A novel approach for the simultaneous recovery of physically plausible trajectories and automatic inference of shape correspondence.
\item Shape transformation using smooth diffeomorphic fields, ensuring topological consistency and non-intersecting trajectories while accommodating  constraints (\eg volume preservation) and physically inspired  priors. 
\item In exchange for providing a simple skeletal structure to augment the source mesh, our method can model rigid (bone) and conformal (soft tissue) deformations within the enclosed volume. These physically inspired soft constraints enhance interpolation quality and help mitigate symmetric ambiguity.
\item Shapes are matched using a Varifold metric, a technique from geometric measure theory, that enables comparison independent of shape fidelity (\eg mesh resolution) or parameterisation. A recently proposed compression technique enhances computational and memory efficiency, enabling effective scaling to densely sampled shapes.
\end{itemize}
\section{Related Work}
\label{sec:related_work}

This section reviews the most relevant methods to the proposed approach, with \Cref{tab:comparison_table} summarising the key attributes of leading techniques.

\begin{table}[t!]
\centering
\setlength{\tabcolsep}{3pt}
\begin{NiceTabular}{@{}lcccc@{}}
\CodeBefore
\rowcolor{lightheadergrey}{3,5,7}
\rowcolor{darkblue}{1}
\Body
\toprule
\multicolumn{1}{c}{\textbf{Method}} & {\textbf{Unsup.}}& {\textbf{No DS}} & {\textbf{No PM}} & {\textbf{Scalable}} \\
\midrule
Div‐Free \cite{eisenberger2019divfree}  & \xmark & \cmark & \cmarkna & \cmark \\ 
Ham. Dyn. \cite{eisenberger2020hamiltonian}  & \xmark & \cmark & \cmarkna & \cmark \\ 
NeuroMorph \cite{eisenberger2021neuromorph} & \cmark & \xmark & \xmark & \xmark \\ 
SMS \cite{cao2024spectral} & \cmark & \xmark & \xmark & \xmark \\ 
ESA \cite{hartman2023elastic} & \cmark & \cmark & \cmark & \xmark \\ 
Ours & \cmark & \cmark & \cmark & \cmark \\
\bottomrule
\end{NiceTabular}
\caption{\textbf{Comparison with leading methods.} Operating in an unsupervised manner (\textbf{\textcolor{blue}{Unsup.}}), doesn't requiring a dataset of samples for training (\textbf{\textcolor{blue}{No DS}}), avoids using a permutation matrix (\textbf{\textcolor{blue}{No PM}}), and scales to high-resolution surfaces (\textbf{\textcolor{blue}{Scalable}}).}
\label{tab:comparison_table}
\vspace{-15pt}
\end{table}

\paragraph{Shape Matching}
\label{sec:shape_matching}\hspace{-0.12em}
\looseness=-1 Shape registration \cite{tam2012registration, van2011survey}, a fundamental problem in computer vision, aims to determine spatial correspondences between pairs of shapes. Use cases range from pose estimation for noisy point clouds to the non-parametric registration of high-resolution medical images.

The Functional Mapping framework, proposed by \mbox{Ovsjanikov} \etal \cite{ovsjanikov2012functional}, efficiently learns mappings between isometrically deformed shapes using the eigenfunctions of the Laplace-Beltrami operator. Dense point-wise correspondences are then recovered via a nearest neighbour search in functional space using Iterative Closest Point (ICP). Due to its simplicity, generalisability and efficiency this framework, has been widely extended to handle non-isometric mappings \cite{eisenberger2020smooth, magnet2022smooth, ren2018continuous, ren2021discrete}, improve matching accuracy \cite{eynard2016coupled, ren2019structured}, and tackle partial \cite{litany2017fully, rodola2017partial} and multi-shape \cite{cohen2020robust, gao2021isometric, huang2020consistent} matching. While deep learning-based approaches \cite{halimi2019unsupervised, litany2017deep, roufosse2019unsupervised} have also been proposed to remove reliance on hand-crafted feature descriptors.

However, two-stage ``match and refine'' approaches often yield sub-optimal dense correspondences. Post-processing techniques \cite{eisenberger2020smooth, melzi2019zoomout, vestner2017product} attempt to improve the final point-wise maps, while Cao \etal \cite{cao2023unsupervised} directly enforce the functional map to be associated with a point-wise map and optimise both simultaneously. Nevertheless, mapping in the spectral domain does not guarantee smooth point-wise correspondences, and solving for a permutation matrix makes it difficult to handle shapes with differing resolutions.

\looseness=-1 Embedding shapes into measure spaces provides a way to address these issues. The understanding of surface geometry in the context of geometric measure theory and calculus of variations has been widely studied in mathematics, leading to the development of parametrisation-invariant shape matching metrics \eg Currents \cite{vaillant2005surface}, \mbox{Varifolds} \cite{charon2013varifold, kaltenmark2017general, buet2017varifold, buet2022weak} and Normal Cycles \cite{roussillon2019representation}. In particular, the Varifold representation has previously been used in the context of human shapes \cite{bauer2021numerical, pierson20223d, hartman2023bare}. 

Despite invariance to parameterisation and robustness to noise, they scale poorly due to the use of dense kernels, limiting their adoption (quadratic scaling). The Keops library \cite{charlier2021kernel,feydy2020fast} has been the de-facto approach to address these issues, using Map-Reduce to minimise memory allocation by decomposing large matrices into sub-problems. This enables parallelised GPU computations via specialised CUDA kernels, but this does not eliminate the calculation of all pairwise distances. In contrast to previous computationally prohibitive approaches to measure-based shape compression \cite{durrleman2009statistical, gori2016parsimonious, hsieh2021metrics}, Paul et al. \cite{paul2024sparse} have recently developed a practical solution utilising Ridge Leverage Score (RLS) sampling, which produces an accurate lower-dimensional representation with substantially improved computational efficiency. Our work leverages this method to enable the scaling to high-resolution target shapes.

\paragraph{Shape Interpolation}
\label{sec:shape_interpolation}
\looseness=-1 Non-rigid interpolation techniques have been extensively 
explored to generate physically plausible morphing paths between the source and target with applications in shape exploration and deformation transfer \cite{sorkine2009interactive, laga2018survey}. Target alignment is typically balanced against adherence to quality metrics \eg distortion minimisation and preservation of local geometric features.

Common approaches include 
posing interpolating trajectories as geodesics in higher-dimensional shape spaces \cite{brandt2016geometric, heeren2012time, heeren2014exploring, wirth2011continuum}, employing local distortion deformation measures like as-rigid-as-possible (ARAP) \cite{sorkine2007rigid} or PriMo \cite{botsch2006primo}. Alternative methods include cage-based deformations \cite{joshi2007harmonic}, continuum mechanics-based approaches \cite{heeren2012time, heeren2014exploring}, and interpolating intrinsic quantities before reconstructing the 3D shape \cite{baek2015isometric}. However, direct vertex offset predictions can lead to geometric artefacts \eg self-intersection, while coarse control cages may overly restrict the deformation process. 

Eisenberger \etal \cite{eisenberger2019divfree, eisenberger2020hamiltonian} formulate the problem as a time-dependent gradient flow, combining divergence-free vector fields for volume preservation with anisotropic ARAP constraints; computational complexity is controlled by learning fields from a data subsample, \eg a few thousand points.

Despite these innovations, many approaches rely on consistent surface parameterisation between source and target, known correspondences a-priori, and computing expensive constraints at each interpolation step.

\paragraph{Combined Approaches}
NeuroMorph \cite{eisenberger2021neuromorph} produces continuous interpolations between meshes while establishing point-to-point correspondence, using two unsupervised neural networks to learn the displacement field and permutation matrix. Cao \etal \cite{cao2024spectral} improved upon this by harmonising spectral and spatial maps, leading to more accurate and smoother point-wise correspondences under large non-isometric deformations. However, these deep learning-based methods are constrained by their reliance on large, diverse shape datasets for time-intensive feature mapping and their use of permutation matrices, which impede scalability and practical applications. Furthermore, modelling deformations as an offset vector applied directly to the vertices doesn't guarantee smooth trajectories, the non-intersection of surfaces or volume preservation by construction. 

Other recent advances have drawn inspiration from geometric measure theory and the Large Deformation \mbox{Diffeomorphic} Metric Mapping (LDDMM) model. Bauer \etal \cite{bauer2021numerical} proposed a framework for surface registration using the Square Root Normal Field (SRNF) pseudo-distance. Building on this, Hartman \etal \cite{hartman2023elastic} employed Varifolds to interpolate between unregistered triangulated surfaces. However, this latter approach not only inherits the scaling limitations associated with Varifolds but also relies on learning complex geodesics defined by intricate partial differential equations (PDEs) to predict the interpolation.
\begin{figure*}[t!]
    \centering
   \includegraphics[width=0.99\linewidth,keepaspectratio]{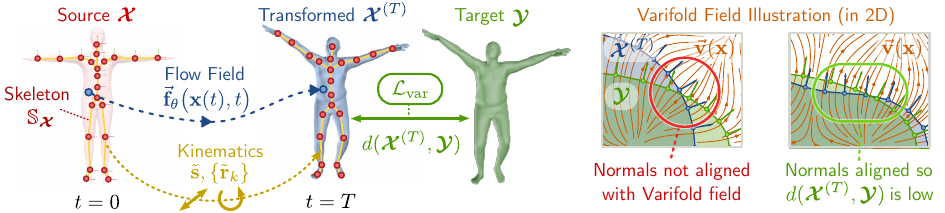}
       \vspace{-5pt}
    \caption{\textbf{Left: Method overview}. We transform the source shape to match the target using a diffeomorphic differential vector field; we also recover the forward kinematic transformation of the source skeleton. Matching is performed with a correspondence-free Varifold metric. 
    \textbf{Right: Varifold field.}~Visualisation of an implicit Varifold field for intuition; computation of $\vec{\mathbf{v}}(\mathbf{x})$ is not required for matching.}
    \label{fig:method_overview}
    \vspace{-10pt}
\end{figure*}

\section{Varifolds in a Nutshell}
\label{sec:varifolds_intro}

Varifold matching~\cite{charon2013varifold} is a technique often used for surface matching in Computational Anatomy~\cite{younes2010shapes} and has its origin in geometric measure theory~\cite{allard1972first}. The goal is to establish a metric for the distance between two shapes without requiring correspondence. We seek to provide a high-level intuition and details of the resulting computation but an in-depth treatment is left to further literature~\cite{paul2024sparse}.

\newcommand{\vf}[0]{\vv{\mathbf{v}}}
\newcommand{\xx}[0]{\mathbf{x}}
\newcommand{\yy}[0]{\mathbf{y}}
\newcommand{\nn}[0]{\mathbf{n}}
\newcommand{\sn}[0]{\hat{\mathbf{n}}}

We consider a vector field $\vf(\xx) : \mathbb{R}^3 \rightarrow \mathbb{R}^3 $ and a surface $\mathbfcal{X}$, and define a scalar measure as the integral of the vector field over the surface, often termed the current, as
\begin{equation}
    \mu_{\mathbfcal{X}}(\vf) := \int_{\mathbfcal{X}} \vf(\xx) \cdot \sn(\xx) \, dS_\mathbfcal{X}(\xx) \ ,
\end{equation}
where we have taken the inner product between the vector field and the unit surface normal $\sn(\xx)$ with $dS_\mathbfcal{X}(\xx)$ as the elemental surface area. We limit the vector fields to be a smooth space $\mathbb{V}$ defined as a vector Reproducing Kernel Hilbert Space (RKHS) with kernel $\kappa(\xx, \xx')$ that defines the spatial correlation across the vector field. This allows us to consider the ``best'' vector field that gives the highest measure (under the dual RKHS norm) as
\begin{equation}
    \| \mu_{\mathbfcal{X}} \|_{\mathbb{V}^*} := \sup_{\vf \in \mathbb{V}, \|\vf\| \leq 1} \big|\, \mu_{\mathbfcal{X}}(\vf) \,\big| \ ; \label{eqn:current_measure}
\end{equation}
intuitively this is the smoothest field that passes through the surface with the vectors most aligned with the surface normals. The reproducing property of the RKHS provides a closed form solution to \cref{eqn:current_measure} through the inner product $\| \mu_{\mathbfcal{X}} \|_{\mathbb{V}^*}^2 = \langle \mu_{\mathbfcal{X}}, \mu_{\mathbfcal{X}} \rangle_{\mathbb{V}^*}$ where the product $\langle \mu_{\mathbfcal{X}}, \mu_{\mathbfcal{Y}} \rangle_{\mathbb{V}^*}$ is
\begin{equation}
     \int_{\mathbfcal{X}} \int_{\mathbfcal{Y}} \kappa(\xx, \yy) \langle \sn_{\mathbfcal{X}}(\xx), \sn_{\mathbfcal{Y}}(\yy) \rangle \, dS_\mathbfcal{X}(\xx) \, dS_\mathbfcal{Y}(\yy) . \label{eqn:current_inner_product}
\end{equation}

\newcommand{\hilbert}[2]{\langle #1, #2 \rangle_{\mathbb{V}^*}}

We can now define a distance between two shapes using norm as $d(\mathbfcal{X}, \mathbfcal{Y}) := \|\mu_{\mathbfcal{X}} - \mu_{\mathbfcal{Y}}\|_{\mathbb{V}^*}^2$ where
\begin{equation}
\begin{aligned}
    d(\mathbfcal{X}, \mathbfcal{Y}) &= \hilbert{\mu_{\mathbfcal{X}} - \mu_{\mathbfcal{Y}}}{\mu_{\mathbfcal{X}} - \mu_{\mathbfcal{Y}}} \\
    &= \hilbert{ \mu_{\mathbfcal{X}} }{ \mu_{\mathbfcal{X}} } 
    - 2 \hilbert{ \mu_{\mathbfcal{X}} }{ \mu_{\mathbfcal{Y}} }
    + \hilbert{ \mu_{\mathbfcal{Y}} }{ \mu_{\mathbfcal{Y}} } \ . \label{eqn:varifold_metric}
\end{aligned}
\end{equation}

The final step to the full Varifold measure is to generalise and include a kernel over the normal vectors as well, rather than the standard Euclidean dot product in \cref{eqn:current_inner_product}, so that we have a spatial kernel $\kappa_{\xx}(\xx, \xx')$ and a normal kernel $\langle \sn_\mathbfcal{X}, \sn_\mathbfcal{Y} \rangle \rightarrow \kappa_{\nn}(\sn_\mathbfcal{X}, \sn_\mathbfcal{Y})$; $\langle \mu_{\mathbfcal{X}}, \mu_{\mathbfcal{Y}} \rangle$ becomes
\begin{equation}
     \int_{\mathbfcal{X}} \int_{\mathbfcal{Y}} \kappa_{\xx}(\xx, \yy) \, \kappa_{\nn}( \sn_{\mathbfcal{X}}(\xx), \sn_{\mathbfcal{Y}}(\yy) ) \, dS_\mathbfcal{X}(\xx) \, dS_\mathbfcal{Y}(\yy) \ . \label{eqn:varifold_inner_product}
\end{equation}

\paragraph{Intuition} 
If we specify Gaussian kernels, that is $\kappa(\xx, \yy) := \exp(-\frac{1}{2 \ell^2} \| \xx - \yy \|^2)$, then intuition for the metric becomes clear. 
The product in \cref{eqn:varifold_inner_product} makes sense as when points $\xx$ and $\yy$ from the two shapes are close, \ie $\| \xx - \yy \|$ is small, then $\kappa_{\xx}(\xx, \yy)$ will be large and we want $\kappa_{\nn}(\sn_{\mathbfcal{X}}(\xx), \sn_{\mathbfcal{Y}}(\yy))$ to be large as well; 
therefore, $\| \sn_{\mathbfcal{X}}(\xx) - \sn_{\mathbfcal{Y}}(\yy) \|$ becomes small and the normals will be equal. Conversely, when points from the different shapes are far apart then $\| \xx - \yy \|$ is large and $\kappa_{\xx}(\xx, \yy) \approx 0$ so the normals can be different; illustrated on the right of \cref{fig:method_overview}. We can pick the lengthscales $\ell_\xx$ and $\ell_\nn$ appropriately for the resolution we want the surfaces to match at (they should be commensurate with the surface discretisation).
The measure is minimised (at 0) when the two surfaces are the same and, importantly, \emph{at no point have we required known correspondences between the two shapes}.
\section{Method}
\label{sec:method}

\newcommand{\skel}[0]{\mathbb{S}_{\mathbfcal{X}}}

\paragraph{Notation}
Shape interpolation continuously deforms a source shape $\mathbfcal{X}$ to match a target shape $\mathbfcal{Y}$. 
Formally, we let $\mathbfcal{X} = \{V_\mathbfcal{X}, N_\mathbfcal{X}, dS_\mathbfcal{X}\}$ consist of a discrete set of vertices on the surface, $\cramped{V_\mathbfcal{X} = \{\mathbf{x}_{i}\}, i \in[1, I]}, \mathbf{x}_{i} \in \mathbb{R}^3$, with associated normal vectors $\cramped{F_\mathbfcal{X} = \{\hat{\mathbf{n}}_{i}\}}$ and surface areas $\cramped{dS_\mathbfcal{X} = \{s_i\}}$; similarly for $\mathbfcal{Y}$. These could be from a triangular mesh but our approach allows for more general surface representations.
In addition, \emph{for the source shape alone}, we provide a simple internal skeleton $\skel = \{ \mathbf{b}_j, e_k \}$ comprising an acyclic graph of internal vertices $\mathbf{b}_j \in \mathbb{R}^3$, $j \in [1, J]$, connected by edges $e_k = (j, j')$, $k \in [1, K]$. 

\paragraph{Overview}
\label{sec:method_overview}
Our method comprises four key components:
\begin{enumerate}
    \item Deformation via a diffeomorphic flow field to guarantee \textbf{preservation of topology}; the vector field is constructed to be divergence free and therefore \textbf{preserves volume}. 
    \item A varifold metric is used to ensure the \textbf{deformed surface matches the target without known correspondence}.
    \item We supply the source with a simple, articulated internal skeleton and then force our field to \textbf{infer the new skeletal pose automatically whilst promoting local rigidity}.
    \item The surface and soft tissue surrounding the skeleton are encouraged to deform in a conformal manner by \textbf{incorporating physically inspired priors}. 
\end{enumerate}
\looseness=-1 This section discusses the components of our model, illustrated in \cref{fig:method_overview}, used to learn the deformation.

\subsection{Diffeomorphic Flow}
\label{sec:diffeomorphic_flow}

The deformation is modelled as a diffeomorphism, represented by a time-varying vector flow field $\FlowField(\mathbf{x}, t) : \mathbb{R}^{3 \times 1} \rightarrow \mathbb{R}^3$, that deforms the surface under an Ordinary Differential Equation (ODE); the dynamics are provided by a neural network as a NeuralODE~\cite{chen2018neural}. The surface is pushed (or ``evolved'') under this flow at all points in a continuous manner, independent of mesh resolution or parameterisation.
As the vector field is continuous and differentiable, the ``streamlines'' of the field will not cross; the surface is guaranteed not to self-intersect and the topology (\eg number of holes) is preserved. Equally, differential properties of the surface or volume (\eg the normal vector to the surface) can also be propagated through the field.

\newcommand{\odeSolve}[0]{\mathsf{ODE_{Solve}}}

\paragraph{Deformation}
A surface position becomes a function of time, $\mathbf{x}(t)$, where we start at $t=0$ on the source shape, $\mathbf{x}(0) \in V_\mathbfcal{X}$, and at some fixed time $t=T$ we want the point to lie on the surface of the target $\mathbfcal{Y}$. Therefore the shape evolves under an initial value ODE
\vspace{-5pt}
\begin{equation}
    \frac{d}{dt} \mathbf{x}(t) = \FlowField( \mathbf{x}(t), t ) , \quad \text{s.t.} \quad \mathbf{x}(t=0) \in V_\mathbfcal{X} \ .
\label{eq:flow_field}
\end{equation}

The estimate of a point $\mathbf{x}^{(0)} := \mathbf{x}(t=0)$ on the source moving to the point  $\mathbf{x}^{(T)} := \mathbf{x}(t=T)$ on the target surface is the solution to the Initial Value Problem (IVP)
\vspace{-5pt}
\begin{equation}
\mathbf{x}^{(T)} = \mathbf{x}^{(0)} + \int_{0}^{T} \FlowField_{\!\theta}( \mathbf{x}(t), t ) \, dt =: \odeSolve \left(  \mathbf{f}_{\theta}, \mathbf{x}_{0}, T \right) \ ,
 \label{eq:ode_flow}
\end{equation}
where we have parameterised the flow field by $\theta$, the neural network weights of a NeuralODE, and defined the $\odeSolve$ function as the result of a numerical solver that approximates the integral to a given tolerance.  

\paragraph{Differential Properties}
If we parameterise our \mbox{NeuralODE} appropriately we can guarantee suitable continuity on the vector field. This allows us to transfer differential properties from source to target. For example, if $\mathcal{J}[\cdot]$ is some linear differential operator that yields the surface normal, \ie $\mathcal{J}[\mathbf{x}] = \mathbf{n}$, we have
\begin{equation}
    \mathbf{n}^{(T)} = \mathbf{n}^{(0)} + \int_{0}^{T} \mathcal{J}\big[ \FlowField_{\!\theta}( \mathbf{x}(t), t ) \big] \, dt \ . \label{eqn:solve_for_normals}
\end{equation}

\paragraph{Volume Preservation}
Many natural shapes, \eg humans and animals, are composed mostly of incompressible tissues that preserve volume under deformation. We incorporate the constraint by learning a divergence-free vector field~\cite{eisenberger2019divfree}. 
The curl of a vector field is always divergence-free, thus we parameterise the vector flow field as 
\begin{equation}
    \FlowField_{\!\theta}(\mathbf{x}, t) := \nabla \times \vv{\mathbf{a}}_{\theta}(\mathbf{x}, t) \ ,
\end{equation}
where $\vv{\mathbf{a}}_{\theta}(\mathbf{x}, t) \in \mathbb{R}^3$ is the output of a neural network and the curl operation is calculated by auto-differentiation.

\paragraph{ARC-Net}
A SIREN~\cite{sitzmann2020implicit} network, with a layer of FINER~\cite{liu2024finer}, is used as to specify the vector field $\vv{\mathbf{a}}_{\theta}(\mathbf{x}, t)$ in our NeuralODE. 
This enables the network to represent fine details in the underlying signal and capture the derivatives of the target function; this addresses the limitations of conventional ReLU-based MLPs. See the supplementary material for details of the precise architecture used. 

\paragraph{ODE Solver}
Our approach is solver-agnostic; however, in recent years, probabilistic solvers have emerged as an efficient framework for integrating uncertainty quantification with inference in dynamical systems~\cite{schober2019probabilistic, kramer2024stable, kramer2022probabilistic, bosch2021calibrated, tronarp2021bayesian, tronarp2019probabilistic}. These solvers utilise a Gaussian Process (GP) function, which offers greater flexibility compared to traditional Runge-Kutta (RK) based polynomial representations, thereby reducing the number of time steps required to solve the dynamics of an underlying ODE function accurately. 
We solve the IVP on a fixed time grid using the \mbox{``KroneckerEK0''} formulation with a single derivative~\cite{kramer2024stable}.

\newcommand{\lossVarifold}[0]{\mathcal{L}_{\text{var}}(\theta)}

\subsection{Varifold Matching}
\label{sec:varifold_matching}

We use a discrete approximation for the integral in \cref{eqn:varifold_inner_product},
\vspace{-5pt}
\begin{equation}
    \langle \mu_{\mathbfcal{X}}, \mu_{\mathbfcal{Y}} \rangle \approx
    \sum_{i=1}^{I_\mathbfcal{X}} \sum_{i'=1}^{I_\mathbfcal{Y}} \kappa_{\xx}(\xx_i, \yy_{i'}) \kappa_{\nn}(\sn_{\mathbfcal{X}_i}, \sn_{\mathbfcal{Y}_{i'}}) s_{\mathbfcal{X}_i} \, s_{\mathbfcal{Y}_{i'}} \ , \label{eqn:discrete_varifold_product}
\end{equation}
to calculate the Varifold metric, \cref{eqn:varifold_metric}, to use as the surface matching loss $\lossVarifold := d(\mathbfcal{X}^{(T)}, \mathbfcal{Y})$. Here, $\mathbfcal{X}^{(T)}$ denotes the set of vertices, normals and differential surface areas that are obtained at time $t=T$ from pushing $\mathbfcal{X}$ through the $\mathsf{ODE_{Solve}}$ in \cref{sec:diffeomorphic_flow}. 

\paragraph{Efficient Scaling}
\label{sec:varifold_compression}
The product computation in \cref{eqn:discrete_varifold_product} is $\mathcal{O}(I_\mathbfcal{X} I_\mathbfcal{Y})$, \ie quadratic in the mesh resolution; this substantial computational burden has previously limited the use of Varifolds. We build on recent work from Paul~\etal~\cite{paul2024sparse} that dramatically reduces the computational cost and allows scaling to high resolution meshes. 
The method compresses each densely sampled surface, $\mathbfcal{S}$, into a lower-dimensional representation, $\mathbfcal{S_{C}}$, via Ridge Leverage Score (RLS) Sampling, to yield an accurate approximation of the Varifold representation; high-resolution shapes can be compressed within seconds while maintaining perceptually lossless quality. 
Post-compression, the same Varifold loss metric, \cref{eqn:varifold_metric}, is computed using the new sample locations, normals, and weights, $\{ V_\mathbfcal{S_{C}}, N_\mathbfcal{S_{C}}, dS_\mathbfcal{S_{C}} \}$, resulting in far smaller kernels and much faster calculation (see~\cref{fig:teaser}).
See the supplementary material for algorithmic details.

\subsection{Skeleton-Driven Transformation}
\label{sec:skeleton_transformation}

\newcommand{\bone}[0]{\mathbf{b}}
\newcommand{\edge}[0]{e}
\newcommand{\trans}[0]{\mathbf{s}}
\newcommand{\quat}[0]{\mathbf{r}}
\newcommand{\pt}[0]{\mathbf{p}_k}

We assume an articulated body has an internal skeletal structure and provide a simple skeleton $\skel = \{\bone_j, \edge_k\}$, of vertices and edges, for the source shape $\mathbfcal{X}$. 
This should deform in a locally rigid manner whilst the surrounding soft tissue and surface deform non-rigidly; we promote this behaviour with an appropriate penalty (loss) on the deformation field, however, we do not know the skeletal pose of the target.
To that end, we simultaneously (i) solve for the forward kinematics of the final skeletal pose (the global translation $\tilde\trans \in \mathbb{R}^3$ and quaternion joint angles $\tilde\quat_k \in \mathbb{Q}$ between each bone), and (ii) penalise the field for any non-rigid deformation across the skeleton. We apply this penalty throughout the interpolation, \ie over $t \in (0, T]$, to ensure the skeleton always remains piecewise rigid.

Solving the forward kinematic chain for the skeleton yields the absolute translation and quaternion for each bone
\begin{equation}
    \{ \trans^{(T)}_k, \quat^{(T)}_k \} = \textsf{FwdKinematics}(\skel, \tilde\trans, \{\tilde\quat_k\}) \ ,
\end{equation}
where the superscript denotes the final pose (at $t = T$).

\newcommand{\boneParams}[0]{\boldsymbol{\alpha}}

\begin{figure}[t]
    \centering
   \includegraphics[width=0.98\linewidth,keepaspectratio]{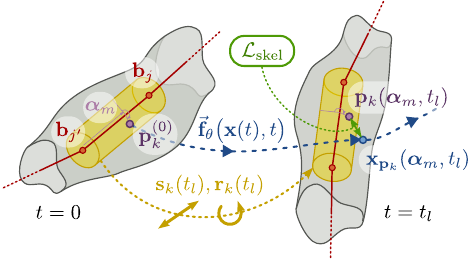}
       \vspace{-6pt}
    \caption{\textbf{Skeleton loss}. We encourage the flow to match an estimated rigid transform for points within the cylindrical ``bones''.}
    \label{fig:skeleton}
\end{figure}

\paragraph{Bone Model}
We model each bone $e_k = (j,j')$ as a cylinder from $\bone_j$ to $\bone_{j'}$ with a radius as a proportion, $\beta$, of the length. 
We sample a random point $\pt^{(0)}(\boneParams)$ in this cylinder where $\boneParams := [\alpha_\tau, \alpha_\rho, \alpha_\psi] \in [0, 1]^2 \times [0, 2\pi]$ are the cylindrical coordinates; $\alpha_\tau$ is scaled to cover the length $0 \rightarrow \| \bone_{j'} - \bone_{j} \|$ and $\alpha_\rho$ the radius $0 \rightarrow \beta \, \| \bone_{j'} - \bone_{j} \|$.
We interpolate the rigid transformation for the point, $\pt^{(0)}(\boneParams)$,
\begin{equation}
    \pt(\boneParams, t) = \trans_k(t) + \quat_k(t) \cdot \pt^{(0)}(\boneParams) \cdot \quat_k^*(t) \
    \label{eq:rigid_tranformation}
\end{equation}
using the SLERP~\cite{shoemake1985animating} transformation with $\trans_k(t) := \frac{t}{T} \trans^{(T)}_k$ and $\quat_k(t) := \exp\big( \frac{t}{T} \ln(\quat^{(T)}_k)\big)$.
\newcommand{\lossBone}[0]{\mathcal{L}_{\text{skel}}(\theta, \tilde\trans, \{ \tilde\quat_k \} )}
Illustrated in~\cref{fig:skeleton}, we compare points in the bones pushed through the ODE solver to their corresponding rigid body estimate, over a set of times $\{t_l\} \in (0,T]$ and samples $\{ \boneParams_m \}$, to provide the loss 
\begin{align}
    \lossBone &:= \sum_{k,l,m}\big\| \pt(\boneParams_m, t_l) - \xx_{\pt}(\boneParams_m, t_l)  \big\|^{2} \ , \nonumber \\
    \xx_{\pt}(\boneParams_m, t_l) &:= \odeSolve( \mathbf{f}_{\theta}, \pt^{(0)}(\boneParams_m), t_l ) \ .
\end{align}

\subsection{Soft Tissue Transformations}
\label{sec:tissue_transformation}

\newcommand{\quatAMAP}[0]{\tilde{\mathbf{q}}}
\newcommand{\ptAMAP}[0]{\mathbf{c}}
\newcommand{\dex}[0]{\hat{\boldsymbol{\delta}}_{\ptAMAP_x}}
\newcommand{\dey}[0]{\hat{\boldsymbol{\delta}}_{\ptAMAP_y}}
\newcommand{\dez}[0]{\hat{\boldsymbol{\delta}}_{\ptAMAP_z}}
\newcommand{\basis}[0]{\mathbf{B}}
\newcommand{\lossSoftBasic}[0]{\mathcal{L}_{\text{soft}}(\theta)}
\newcommand{\lossSoft}[0]{\mathcal{L}_{\text{soft}}(\theta, \phi)}
\newcommand{\qnet}[0]{\tilde{\mathbf{q}}}
\newcommand{\lossSurf}[0]{\mathcal{L}_{\text{surf}}(\theta, \phi)}

\looseness=-4 The surface and soft tissues will deform non-rigidly around the bones (under the constraint of volume preservation); however, they are not a free-flowing fluid and will resist non-rigid deformation with shear or strain energies. We model this as a physically inspired penalty that resists arbitrary transformations; this is related to existing physical based priors such as ``as-rigid-as-possible''~\cite{sorkine2007rigid} or ``as-conformal-as-possible''~\cite{yoshiyasu2014conformal}. 

Inspired by continuum mechanics~\cite{vaxman2015conformal}, we consider how the local basis vectors around a point (\ie an elemental cube) deform. 
We sample a point $\ptAMAP^{(0)}$ in the soft tissue and consider three unit differential axis-aligned basis vectors $\dex^{(0)}, \dey^{(0)}, \dez^{(0)}$ centred on $\ptAMAP^{(0)}$. This evolves as 
\begin{equation}
    \ptAMAP(t) := \odeSolve \big( \mathbf{f}_{\theta}, \ptAMAP^{(0)}, t \big) , \; t \in (0, T] \ ,
\end{equation}
with $\dex(t), \dey(t), \dez(t)$ found using the ODE solver with the Jacobian of the neural vector field as for the surface normals in \cref{eqn:solve_for_normals}.
The best rigid estimate minimises
\begin{equation}
    \min_{\quatAMAP \in \mathbb{Q}} \| \basis_{\ptAMAP}^{(0)} - \quatAMAP^* \cdot \basis_{\ptAMAP}(t) \cdot \quatAMAP \|^2 \ , \label{eqn:basic_soft_loss}
\end{equation}
where $\basis_{\ptAMAP}(t) := [\dex(t), \dey(t), \dez(t)]$ is the transformed basis and $\basis_{\ptAMAP}^{(0)} := [\dex^{(0)}, \dey^{(0)}, \dez^{(0)}] \equiv \mathbf{I}$ is the identity basis.

\paragraph{Q-NET} 
Rather than have to solve for the optimal quaternion $\quatAMAP$ in \cref{eqn:basic_soft_loss} for every point, we use a network to learn the optimal quaternion as a (smooth) function of space and time $\qnet_\phi(\xx, t) : \mathbb{R}^{3 \times 1} \rightarrow \mathbb{Q}$.
This comprises as small specialist MLP, parameterised by weights $\phi$. 
Specifically, we use the method of Peretroukhin~\etal~\cite{peretroukhin2020smooth} who represent rotations through a symmetric matrix that defines a Bingham distribution over unit quaternions.

\begin{figure}[t]
    \centering
   \includegraphics[width=0.98\linewidth,keepaspectratio]{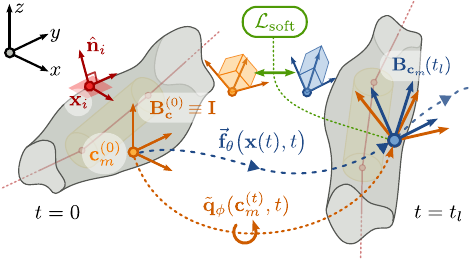}
       \vspace{-6pt}
    \caption{\textbf{Soft tissue loss}. We penalise arbitrary deformation of the soft tissue (and surface) using physically inspired priors to minimise shear/strain energy throughout the flow.}
    \vspace{-14.0pt}
    \label{fig:soft_tissue}
\end{figure}

\paragraph{Tissue/Surface Loss}
The soft tissue loss (\cref{fig:soft_tissue}) is
\vspace{-10pt}
\begin{multline}
    \lossSoft := \sum_{l,m} \| \tilde{\basis}_{m,l} - \mathbf{I} \|^2 + \frac{1}{3} \sum_{n=1}^{3} \big( \big\| [\tilde{\basis}_{m,l}]_n \big\| - 1 \big)^2 , \label{eqn:soft_loss} \\
    \tilde{\basis}_{m,l} := \qnet_\phi^*(\ptAMAP_m^{(t_l)}, t_l) \cdot \basis_{\ptAMAP_m}(t_l) \cdot \qnet_\phi(\ptAMAP_m^{(t_l)}, t_l) \ .
\end{multline}
We evaluate over time samples $\{t_l\}$ for points randomly sampled in the soft tissue $\{ \ptAMAP^{(0)}_m \}$ as well as points on the surface $\{\xx_i\}$.
For surface points, $\lossSurf$ is the same as $\lossSoft$ with $\{ \ptAMAP^{(0)}_m \} \rightarrow\{\xx_i\}$ but we align the first component of the basis vector with the surface normal and only consider the transformation in the 2D basis of the surface.

\subsection{Full Loss Function}
\label{sec:loss_function}

We jointly optimise the parameters of the ARC-Net, $\theta$, defining the time-varying vector flow field, the Q-Net, $\phi$, predicting rotation of conformal samples, and the translation $\tilde\trans$ and joint rotations $\{ \tilde\quat_k \}$ of the forward kinematic skeleton. The full loss is given as
\vspace{-5pt}
\begin{multline}
    \mathcal{L}_{\text{full}}(\theta, \phi, \tilde\trans, \{ \tilde\quat_k \}) := \lossVarifold + \lambda_{1} \, \lossBone \\ + \lambda_{2} \, \lossSoft + \lambda_{3} \, \lossSurf \ , \label{eqn:full_loss}
\end{multline}
where $\lambda_{1}, \lambda_{2}, \lambda_{3} \geq 0$, are weighting parameters for each term; details are provided in the supplemental material. 

\paragraph{Implementation Details}
\looseness=-1 We use the JAX framework \cite{jax2018github} and the Equinox library \cite{kidger2021equinox} for building the neural network architecture. In addition, the ProbDiffEq \cite{kramer2024implementing} library provides the probabilistic ODE solver. The loss function is optimised using VectorAdam~\cite{ling2022vectoradam}, which adapts the basic Adam algorithm to be rotation-equivariant by accounting for the vector structure of optimisation variables. Additional details can be found in the supplementary material. 

\begin{table*}[b!]
\vspace{-5pt}
\begin{adjustbox}{max width=\textwidth}
\centering
\small
\setlength{\tabcolsep}{3pt}
\newcommand{\subhead}[1]{\multicolumn{3}{c}{\begin{tabular}{@{}c@{}} #1 \end{tabular}}}
\begin{NiceTabular}{@{}l *{9}{r@{\,}c@{\,}l} @{}}[colortbl-like]
\CodeBefore
\columncolor{lightheadergrey}{5-7,14-16,23-25}
\rowcolor{darkblue}{1-2}
\rowcolor{lightheadergrey}{4,6,9}
\Body
\toprule
\multicolumn{1}{c}{\multirow{2}{*}{\textbf{Method}}} & \multicolumn{9}{c}{\textbf{MANO}} & \multicolumn{9}{c}{\textbf{DFAUST}} & \multicolumn{9}{c}{\textbf{SMAL}} \\
\cmidrule(lr){2-10} \cmidrule(lr){11-19} \cmidrule(lr){20-28}
& \subhead{\textbf{Geodesic}}$\uparrow$ & \subhead{\textbf{Chamfer}}$\uparrow$ & \subhead{\textbf{Conformal}}$\uparrow$ 
& \subhead{\textbf{Geodesic}}$\uparrow$ & \subhead{\textbf{Chamfer}}$\uparrow$ & \subhead{\textbf{Conformal}}$\uparrow$ 
& \subhead{\textbf{Geodesic}}$\uparrow$ & \subhead{\textbf{Chamfer}}$\uparrow$ & \subhead{\textbf{Conformal}}$\uparrow$ \\
\midrule
Ours & \textbf{0.89} & $\pm$ & 0.17 & {\textbf{0.83}} & $\pm$ & 0.09 & \textbf{0.80} & $\pm$ & 0.13 & \textbf{0.95} & $\pm$ & 0.09 & \textbf{0.89} & $\pm$ & 0.06 & \textbf{0.71} & $\pm$ & 0.17 & \textbf{0.93} & $\pm$ & 0.09 & \textbf{0.87} & $\pm$ & 0.05 & \textbf{0.69} & $\pm$ & 0.08 \\
SMS \cite{cao2024spectral} & 0.83 & $\pm$ & 0.23 & 0.80 & $\pm$ & 0.11 & 0.58 & $\pm$ & 0.17 & 0.88 & $\pm$ & 0.31 & 0.85 & $\pm$ & 0.11 & 0.65 & $\pm$ & 0.26 & 0.82 & $\pm$ & 0.12 & 0.79 & $\pm$ & 0.05 & 0.56 & $\pm$ & 0.12 \\
ESA \cite{hartman2023elastic} & 0.79 & $\pm$ & 0.35 & 0.61 & $\pm$ & 0.41 & 0.63 & $\pm$ & 0.24 & 0.76 & $\pm$ & 0.49 & 0.69 & $\pm$ & 0.33 & 0.70 & $\pm$ & 0.23 & 0.82 & $\pm$ & 0.18 & 0.66 & $\pm$ & 0.16 & 0.57 & $\pm$ & 0.15 \\
Ham.~Dyn. \cite{eisenberger2020hamiltonian} & \multicolumn{3}{c}{n/a} & 0.82 & $\pm$ & 0.05 & 0.62 & $\pm$ & 0.23 & \multicolumn{3}{c}{n/a} & 0.82 & $\pm$ & 0.25 & 0.68 & $\pm$ & 0.28 & \multicolumn{3}{c}{n/a} & 0.76 & $\pm$ & 0.21 & 0.59 & $\pm$ & 0.16 \\
Div.~Free \cite{eisenberger2019divfree} & \multicolumn{3}{c}{n/a} & 0.50 & $\pm$ & 0.28 & 0.53 & $\pm$ & 0.28 & \multicolumn{3}{c}{n/a} & 0.62 & $\pm$ & 0.23 & 0.63 & $\pm$ & 0.28 & \multicolumn{3}{c}{n/a} & 0.58 & $\pm$ & 0.13& 0.49 & $\pm$ & 0.19 \\
\midrule
\midrule
\textsc{Ours + \xmark{} Soft \xmark{} Skel} & 0.84 & $\pm$ & 0.17 & 0.77 & $\pm$ & 0.19 & 0.22 & $\pm$ & 0.32 & 0.92 & $\pm$ & 0.13 & 0.81 & $\pm$ & 0.09 & 0.33 & $\pm$ & 0.37 & 0.88 & $\pm$ & 0.18 & 0.77 & $\pm$ & 0.06 & 0.23 & $\pm$ & 0.24 \\
\textsc{Ours + \xmark{} Soft \cmark{} Skel} & 0.85 & $\pm$ & 0.20 & 0.78 & $\pm$ & 0.12 & 0.21 & $\pm$ & 0.33 & 0.91 & $\pm$ & 0.18 & 0.82 & $\pm$ & 0.10 & 0.34 & $\pm$ & 0.38 & 0.90 & $\pm$ & 0.15 & 0.77 & $\pm$ & 0.06 &  0.23 & $\pm$ & 0.25 \\
\bottomrule
\end{NiceTabular}
\end{adjustbox}
\caption{\textbf{Area Under the Curve (AUC) metrics across all datasets.} AUC metrics are calculated with a threshold of 0.20, 0.1 and 0.15 for Geodesic, Chamfer and Conformal metrics respectively. The top shows comparisons with other methods, the bottom shows an ablation study with the soft tissue (\textsc{Soft}) and skeleton terms (\textsc{Skel}) removed.}
\label{tab:combined_comparison_and_ablation}
\vspace{-15pt}
\end{table*}

\section{Experiments and Discussion}
\label{sec:experimental_results}

We evaluate against four existing approaches: SMS~\cite{cao2024spectral} (dense correspondence and interpolation, requires training data); ESA~\cite{hartman2023elastic} (Varifold-based interpolation without correspondence);  Hamiltonian Dynamics~\cite{eisenberger2020hamiltonian} (high-quality interpolations, requires known dense correspondence); and Divergence Free~\cite{eisenberger2019divfree} (volume preservation via divergence-free Eigen-basis). 
We use three standard datasets: DFAUST~\cite{bogo2017dynamic}, MANO~\cite{romero2017embodied}, and SMAL~\cite{zuffi20173d}; SMAL samples are from a modified version of the dataset \cite{huang2021arapreg}. 
To standardise across methods, all examples are normalised to fit a length-one cube. 
We use K-medoids \cite{park2009simple} to select a diverse range of poses from each dataset and sample random pairs. 
We pick 80 shapes for training (for SMS only) and 20 for testing. For DFAUST, we select random pairs of poses across multiple subjects but ensure a pairing is of the same individual. We remove dataset mesh connectivity bias by remeshing each shape using ACVD~\cite{valette2008generic}.

\paragraph{Metrics}
We quantify performance using three standard metrics via Area Under the Curve (AUC) values: \emph{Geodesic Distance} between predicted and ground truth correspondences (assesses point-to-point mapping accuracy); \emph{Chamfer Distance} between predicted and target surfaces (evaluates overall shape similarity); and \emph{Conformal Distortion} resulting from the interpolation (measures preservation of local geometric properties).
Further details, comprehensive results and analyses are in the supplementary material.

\begin{figure}[t!]
\centering
\includegraphics[width=\linewidth, keepaspectratio]{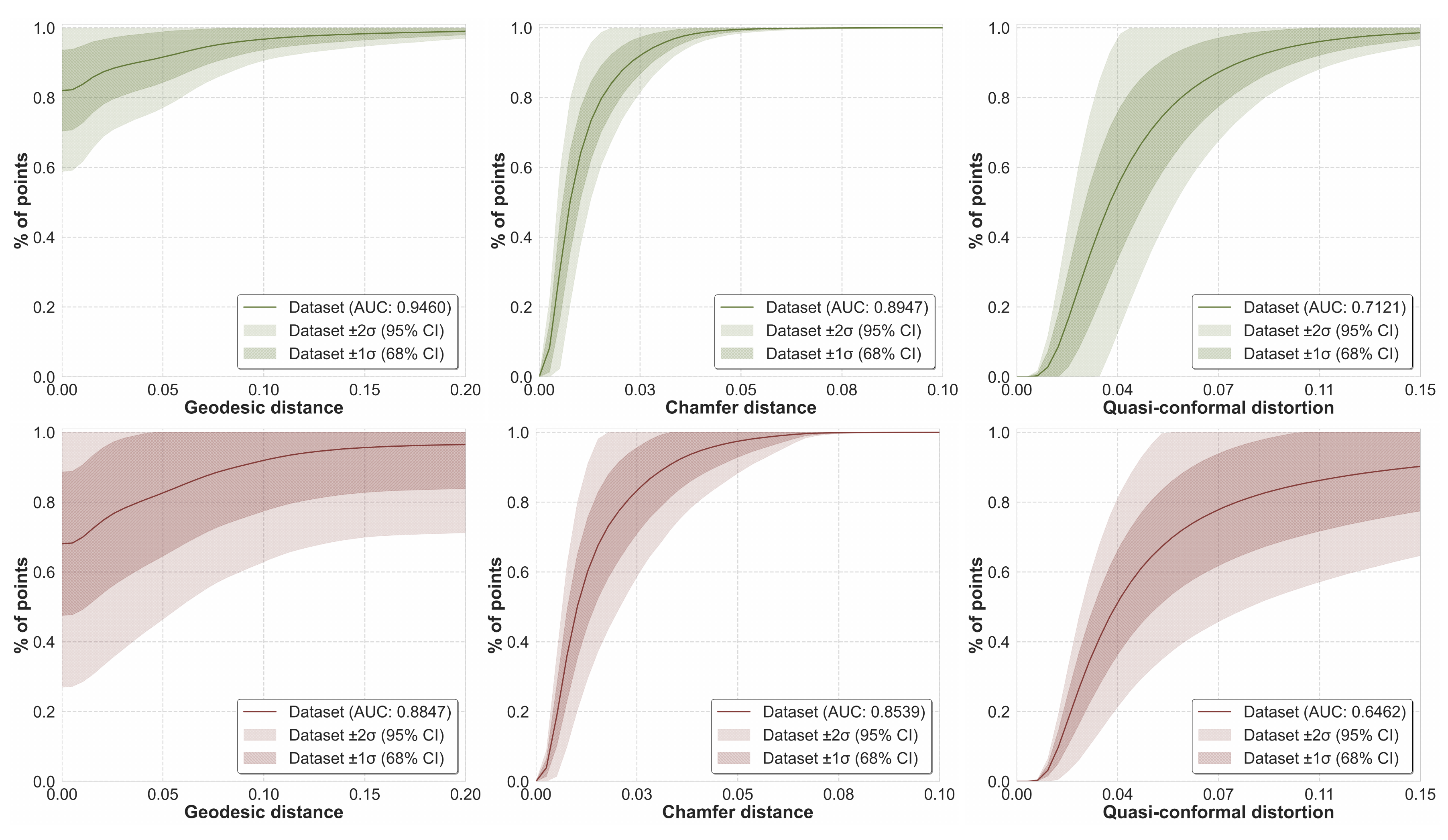}
\caption{\textbf{Interpolation results for DFAUST vs SMS~\protect\cite{cao2024spectral}.} Mean and confidence intervals for the three metrics are shown; top row has our results and bottom show SMS.
Our method improves across all metrics and also has narrower error bars indicating more consistent performance. Please zoom for details.
}
\vspace{-15pt}
\label{fig:graph_results}
\end{figure}

\paragraph{Shape Interpolation}
\Cref{fig:graph_results} shows the interpolation results for our method against the state-of-the-art SMS method for DFAUST showing the mean and confidence intervals for all metrics. Our method improves under all metrics with notably lower variance indicating more consistent performance as well.
Equivalent plots for other comparisons are available in the supplemental material with the results summarised as AUC values for all comparators and datasets in the top half of \cref{tab:combined_comparison_and_ablation}. 
Our approach obtains superior dense correspondence recovery (geodesic distance) without sacrificing the quality of surface fit to the target mesh (chamfer distance) and whilst still conferring the superior theoretical properties of a diffeomorphic transformation. Equally, the results on local surface geometry are improved (conformal distortion).
Qualitative interpolation results against SMS on SMAL are shown in the top of \cref{fig:teaser} where we see an example with significant non-rigid deformation that SMS is unable to model.

\paragraph{Ablation Study}
Since the comparator methods do not require the provision of a source skeleton, \cref{tab:combined_comparison_and_ablation} also includes comparisons showing the effects of omitting the skeleton and soft constraints. The big reduction in the conformal metric without the soft constraints strongly suggests that our physical differential priors are a key factor in recovering good local surface geometry. The benefit of the skeleton (without the soft constraints) is harder to infer from the reductive numbers in the table but centres on robustness; without the skeleton the Varifold can fall into local minima and fail to capture correct correspondences in challenging circumstances (\eg fingers).

\paragraph{Surface Fidelity and Scaling}
The bottom of \cref{fig:teaser} shows that we can match surfaces with vastly different resolutions (\eg 7k matched to 500k vertices). Varifold compression allows us to perform this in a computationally efficient manner; we can see the Chamfer error results for 10k compressed mesh are visually indistinguishable from the full resolution but obtained at a huge reduction in computational (and memory) cost (plot on the right of \cref{fig:teaser}).

\begin{figure*}[t!]
    \centering
  \includegraphics[width=0.99\linewidth,keepaspectratio]{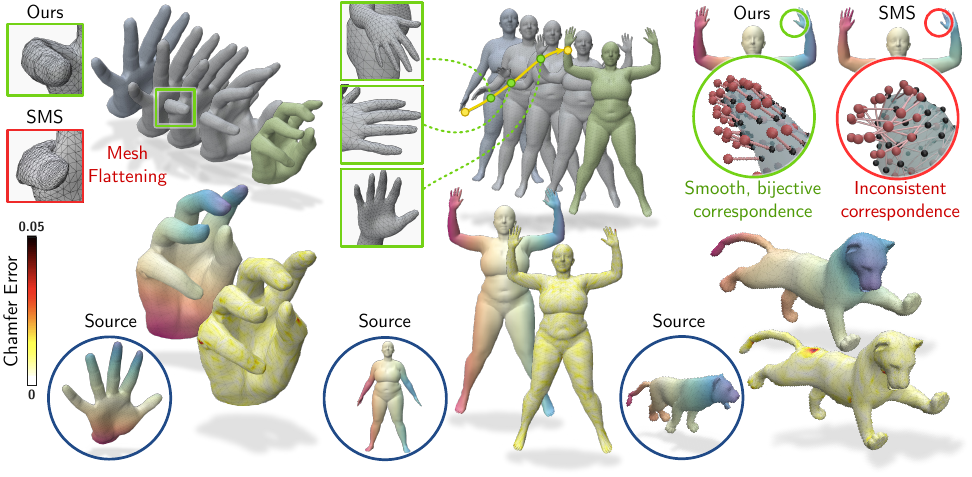}
\vspace{-10pt}
    \caption{\textbf{Challenging examples from each dataset.}
    \textbf{Top Left: Sample interpolations.} Even under large deformations, we do not suffer from flattening or ``rubber hand'' syndrome due to the articulated structure and the soft tissue priors.
    \textbf{Top Right: Correspondence detail.} We guarantee smooth, bijective correspondences by construction in contrast to inconsistent correspondences from permutation matrices.
    \textbf{Bottom: Correspondence and chamfer distances.} Accurate correspondences are recovered with low chamfer errors.
    }
\vspace{-10pt}
    \label{fig:sample_interpolations_and_chamfer_examples}
\end{figure*}

\paragraph{Challenging Examples}
In \cref{fig:sample_interpolations_and_chamfer_examples} we visualise the high quality interpolation path, correspondences and low chamfer errors on the most challenging examples (test set pairs that contain the largest average deformation) from each dataset (the SMAL interpolation is in \cref{fig:teaser}).
\looseness=-3 The offset correspondence visualisation (top right) shows the smooth bijjective correspondences guaranteed by construction from our method whereas permutation matrices can yield inconsistent correspondences (\eg SMS).

\paragraph{Timing}
Our method, without Varifold compression, yields runtimes comparable to ESA, requiring minutes to compute high-quality trajectories and dense correspondences between dataset samples. In contrast, SMS requires a minimum of 24 hours compute on high-performance GPU hardware to train its deep Functional Mapping network, though subsequent inference is rapid (under one minute). The Hamiltonian and Div-Free methods offer the fastest runtime for interpolation however they work from known correspondences making direct comparisons unsuitable.
We note that SMS cannot train on high resolution meshes (due to the permutation matrix) whereas the Varifold compression removes this barrier for our approach; ESA has costly dense pairwise computations that also scale poorly.

\begin{figure}[b!]
 \vspace{-15.0pt}
    \centering
   \includegraphics[width=\linewidth]{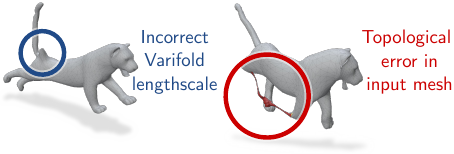}
    \vspace{-17.5pt}
    \caption{\textbf{Failure Cases.} Illustrations of failures when modelling assumptions are not met; \eg the Varifold lengthscale is too small for mesh resolution or the input is topologically inconsistent.}
    \label{fig:failure_cases}
\end{figure}

\paragraph{Failure Cases}
Whilst we found our method to be more robust than the comparator methods in general, there are potential failure cases as shown in \cref{fig:failure_cases}. If the Varifold lengthscale is set incorrectly (given the mesh resolution) then artefacts can be introduced (as on the left). We cannot recover from topology errors in the source or target mesh (since this violates the diffeomorphic deformation; we show an example where the initial mesh had legs self-intersecting.

\section{Conclusion}
We have presented a novel unsupervised framework that learns realistic interpolation trajectories between different 3D shapes and computes accurate shape correspondences in an automatic manner and without relying on any prior knowledge. The effectiveness of the proposed approach was demonstrated using qualitative examples and quantitative analysis in various experiments, including human body shape and pose data as well as human hand scans.

Overall, we believe that our method will be a valuable contribution for bringing shape matching to practical applications in challenging real-world settings especially the ability to scale to high resolution surfaces and operate under arbitrary surface representations. 

As future work, we plan to explore the initialisation and optimisation of the source skeleton, particularly using automatic skeleton generation techniques \cite{xu2020rignet, ma2023tarig, novakovic2024learning}, and examine the impact of different skeletal structures.

\FloatBarrier

\newpage

{
    \small
    \bibliographystyle{ieeenat_fullname}
    \bibliography{main}
}

\clearpage
\setcounter{section}{0}
\renewcommand{\thesection}{S\arabic{section}}
\maketitlesupplementary
\section{Implementation Details}
\label{sec:sup_implementation_details}

In this section, we provide implementation details that were omitted from the main text due to space constraints. 

\subsection{ARC-Net Architecture}
\label{sec:arc_net_architecture}

\begin{figure}[h!]
\vspace{-10pt}
    \centering
   \includegraphics[width=\linewidth]{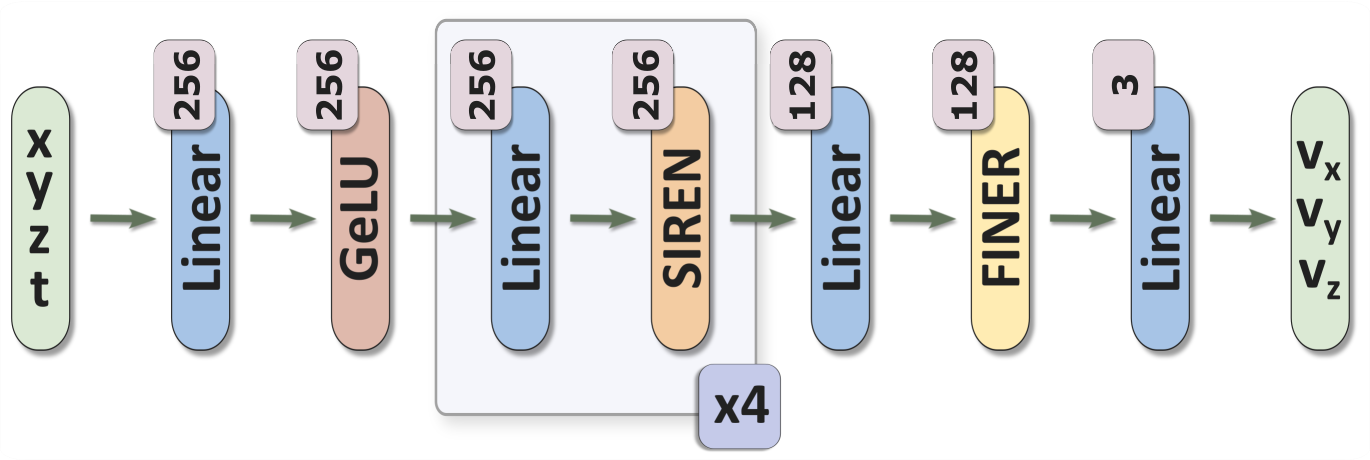}
       \vspace{-10pt}
    \caption{\textbf{Overview of the ARC-Net Architecture.}}
    \vspace{-5pt}
    \label{fig:flow_net_architecture}
\end{figure}

The ARC-Net introduced in \Cref{sec:diffeomorphic_flow} comprises an MLP with 4 SIREN layers and one FINER layer, with widths of 256 and 128 respectively, as shown in \cref{fig:flow_net_architecture}. SIREN \cite{sitzmann2020implicit} uses a sine as a periodic activation function such that $\mathbf{z}_{i}: \mathbb{R}^{M_i} \rightarrow \mathbb{R}^{N_i}$ is the $i$-th layer of the network is defined as,
\begin{equation}
\begin{aligned}
\mathbf{z}_i &= \sin(\omega_0( \mathbf{W}_i \mathbf{z}_{i-1} + \mathbf{b}_i)) \ , 
\end{aligned}
\end{equation}
where $\mathbf{z}_{i-1}$ denotes the output of layer $i-1$ and $\omega_0$ is a user defined parameter for controlling the frequency. For ARC-NET, we use $\omega_{0} = 4.0$.

However, the standard formulation exhibits a well-documented spectral bias, wherein the network preferentially learns low-frequency components of the signal. While this bias can be advantageous for learning smooth flow fields, it potentially limits the network's ability to handle fine grain deviations required to handle intricate surface details on target surfaces. Thus, to address this limitation, we append a FINER layer \cite{liu2024finer} as the final layer, which replaces SIREN's conventional sine activation with a variable-periodic activation function,
\begin{equation}
\begin{aligned}
\mathbf{z}_{i} &= \sin ( \omega_{0} \alpha_{i} ( \mathbf{W}_{i} \mathbf{z}_{i-1} + \mathbf{b}_{i} )) \ , \\ 
\text{where} \ \alpha_{i} &= | \mathbf{W}_{i} \mathbf{z}_{i-1} + \mathbf{b}_{i} | + 1
\end{aligned}
\tag{3}
\end{equation}

\subsection{Q-Net Architecture}

Rotations in three-dimensional space can be represented through various mathematical formulations, including Euler angles, rotation matrices, axis-angle vectors, and unit quaternions. Our model employs unit quaternions due to their compact representation and straightforward geometric and algebraic properties. However, despite these advantages, it is known that naively attempting to learn large rotations via a neural network is problematic due to a critical limitation in the smoothness characteristics of unit quaternions, which can impede the network's ability to accurately learn the correct rotation.

Thus, the Q-Net discussed in \Cref{sec:tissue_transformation} 
leverages the work of Peretroukhin~\etal~\cite{peretroukhin2020smooth}. The network architecture comprises three fully connected layers, each with a width of 128 neurons using Tanh activation functions. Provided with a 4d input, $(x,y,z,t)$, the output of the MLP produces a 10-dimension output which is used to construct the following $4 \times 4$ symmetric matrix, 
\begin{equation}
\mathbf{A}(\theta) = 
\begin{bmatrix}
\theta_1 & \theta_2 & \theta_3 & \theta_4 \\
\theta_2 & \theta_5 & \theta_6 & \theta_7 \\
\theta_3 & \theta_6 & \theta_8 & \theta_9 \\
\theta_4 & \theta_7 & \theta_9 & \theta_{10}
\end{bmatrix},
\end{equation}
This represents the set of real symmetric $4 \times 4$ matrices with a simple (i.e., non-repeated) minimum eigenvalue:
\begin{equation}
\mathbf{A} \in \mathbb{S}^4 : \lambda_1(\mathbf{A}) \neq \lambda_2(\mathbf{A}),
\tag{1}
\end{equation}
where $\lambda_i$ are the eigenvalues of $\mathbf{A}$ arranged such that $\lambda_1 \leq \lambda_2 \leq \lambda_3 \leq \lambda_4$, and $\mathbb{S}^n \triangleq \{\mathbf{A} \in \mathbb{R}^{n\times n} : \mathbf{A} = \mathbf{A}^\top\}$. 
\begin{figure}[h!]
\vspace{-10pt}
    \centering
   \includegraphics[width=0.75\linewidth]{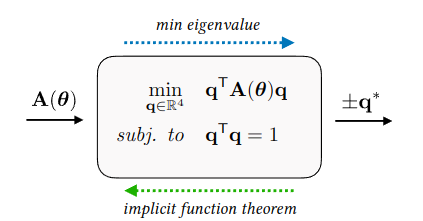}
    \caption{\textbf{QCQP layer} - Image Credit \cite{peretroukhin2020smooth}}
    \vspace{-5pt}
    \label{fig:qcqp_layer}
\end{figure}

$\mathbf{A}(\theta)$ is mapped to a unique rotation through a differentiable QCQP layer, illustrated in \Cref{fig:qcqp_layer}. This layer predicts a quaternion as a solution derived from the minimum-eigenspace of $\mathbf{A}$ and the implicit function theorem allows for an analytic gradient to be computed for use as part of back-propagation, 
\begin{equation}
\frac{\partial\mathbf{q}^*}{\partial\text{vec}(\mathbf{A})} = \mathbf{q}^* \otimes (\lambda_1\mathbf{I} - \mathbf{A})^\dagger \ ,
\tag{9}
\end{equation}
where $(\cdot)^\dagger$ denotes the Moore-Penrose pseudo-inverse, $\otimes$ is the Kronecker product, and $\mathbf{I}$ refers to the identity matrix.

\subsection{Skeleton Parameterisation}

\begin{figure*}[t!]
    \centering
   \includegraphics[width=\linewidth]{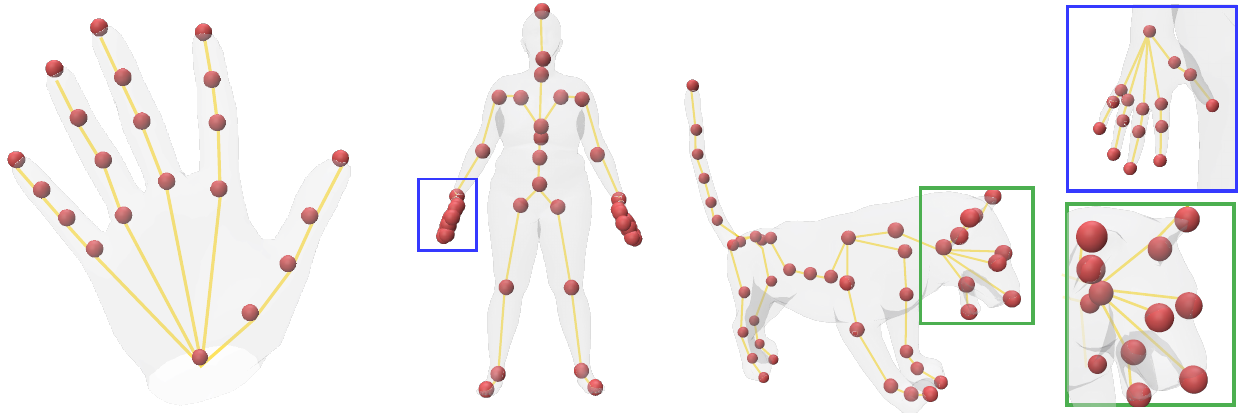}
    \caption{\textbf{The simple skeletons, $\skel$, based upon skeleton provided with each dataset, used to augment the source shapes in our method.} \textbf{Left: MANO.} consisting of 21 joints and 21 bones. \textbf{Centre: DFAUST.} consisting of 53 joints and 52 bones. \textbf{Right: SMAL.} consisting of 44 joints and 43 bones.}
    \label{fig:arc_skeletons}
\end{figure*}

We utilise the skeleton provided with each dataset with a minor adjustment. To enhance realism, we extend the existing skeleton by adding bones at the extremities; specifically in the fingers, feet, and paws. This augmentation results in a more anatomically accurate representation, more closely mimicking real-life skeletal structures as shown in \cref{fig:arc_skeletons}.

The exact regions defined as rigid / soft tissue and the number of samples used are summarised in \Cref{tab:rigid_soft_tissue_sampling_table}. In general, for each bone, we use 50 samples for the bone, $\left\{ \alpha_{m} \right\}$, and the soft tissue, $\left\{ c_{m} \right\}$, components, while 500 samples are used for the surface points, $\left\{ x_{i} \right\}$. All of which are randomly resampled in every epoch. We use a radius of 10\% of the length of each bone and between 10\% and 25\% depending on the dataset for the soft tissue region. Human fingers are relative to the rest of a human body long and narrow, thus for DFAUST dataset they require significantly smaller regions to stay within the surface.

\begin{table}[b!]
\centering
\setlength{\tabcolsep}{3pt}
\begin{NiceTabular}{@{}lcccc@{}}
\CodeBefore
\rowcolor{lightheadergrey}{4,6}
\rowcolor{darkblue}{1-2}
\Body
\toprule
\multirow{2}{*}{\textbf{Dataset}} & \multicolumn{2}{c}{\textbf{Rigid} $\left\{ \alpha_{m} \right\}$}  & \multicolumn{2}{c}{\textbf{Soft Tissue} $\left\{ c_{m} \right\}$} \\
\cmidrule(lr){2-3} \cmidrule(lr){4-5}
& \textbf{Radius}  & \textbf{\# Samples} & \textbf{Radius}  & \textbf{\# Samples} \\
\midrule
\textbf{MANO}  & 0.1 & 50 & 0.25 & 50 \\ 
\textbf{DFAUST} (Body) & 0.1 & 50 & 0.25 & 25 \\ 
\textbf{DFAUST} (Hands) & 0.01 & 50 & 0.02 & 25 \\ 
\textbf{SMAL} & 0.1 & 50 & 0.15 & 50 \\ 
\bottomrule
\end{NiceTabular}
\caption{\textbf{Parameters for Rigid \& Soft Tissue Sampling:} The radii as defined in terms of the percentage of the bone length and used for both the bone and soft tissue regions.}
\label{tab:rigid_soft_tissue_sampling_table}
\vspace{-15pt}
\end{table}

\subsection{Training details}

In this section, we provide further details of the training procedure and parameters used to fit our model to the various datasets tested, a summary which is provided in \Cref{tab:combined_opt_parameters}. 

As discussed in the main text, the loss function, \Cref{eqn:full_loss}, is optimised using the VectorAdam \cite{ling2022vectoradam} algorithm and the ODE modelling of the flow field is solved using a probabilistic ODE solver, specifically the Kronecker EK0 formulation with a single derivative operating with a smoother strategy. We use a variable learning rate, which is controlled via a warmup cosine decay schedule, in which 50 epochs are assigned to the warm up stage and the initial and final rates for different datasets are shown in the aforementioned table.

The model is optimised in two stages; main and fine-tuning (FT). During the main phase, after 1k, 2k and 3k epochs, the length scales of the Varifold kernels are adjusted in a coarse to fine manner, after which it is held constant for the remainder of the optimisation. Although many parameters vary slightly depending on the dataset, the weightings of $\mathcal{L}_{\text{bone}}$, $\mathcal{L}_{\text{soft}}$, and $\mathcal{L}_{\text{surf}}$, represented by $\boldsymbol{\lambda_{1}}$, $\boldsymbol{\lambda_{2}}$, and $\boldsymbol{\lambda_{3}}$ respectively, are consistent across all datasets. 

Following the completion of the main stage, the values of $\boldsymbol{\lambda_{1}}$ and $\boldsymbol{\lambda_{2}}$ are increased (all other parameters are held constant), and the network is trained for an additional 2k epochs with these increased values. This additional fine-tuning step was found to improve the quality of the interpolation, both qualitatively and quantitatively (via the conformal metric). Starting initially with these higher values was problematic, as they place a high cost on any initial deformation of the source surface towards the target. 

\begin{table*}[t!]
\centering
\small
\setlength{\tabcolsep}{4pt}
\renewcommand{\arraystretch}{0.85}
\begin{minipage}{.75\textwidth}
\begin{NiceTabular}{l cc cc c cccccccc}[colortbl-like]
\CodeBefore
\columncolor{lightheadergrey}{3,5,7,9,11,13}
  \rowcolor{darkblue}{1-2}
  \rowcolor{lightheadergrey}{4}
\Body
\toprule
\multirow{2}{*}{\textbf{Dataset}} & \multicolumn{2}{c}{\textbf{Epochs}} & \multicolumn{2}{c}{\textbf{LR}} & \textbf{ODE} & \multicolumn{2}{c}{\textbf{Initial}} & \multicolumn{2}{c}{\textbf{Epoch 1k}} & \multicolumn{2}{c}{\textbf{Epoch 2k}} & \multicolumn{2}{c}{\textbf{Epoch 3k}} \\
\cmidrule(lr){2-3} \cmidrule(lr){4-5} \cmidrule(lr){7-8} \cmidrule(lr){9-10} \cmidrule(lr){11-12} \cmidrule(lr){13-14} 
& \textbf{Main} & \textbf{FT} & \textbf{Init} & \textbf{Final} & \textbf{Steps} & $\ell_{\kappa_x}$ & $\ell_{\kappa_n}$ & $\ell_{\kappa_x}$ & $\ell_{\kappa_n}$ & $\ell_{\kappa_x}$ & $\ell_{\kappa_n}$ & $\ell_{\kappa_x}$ & $\ell_{\kappa_n}$ \\
\midrule
\textbf{MANO} & 4k & 2k & 1e-2 & 1e-3 & 10 & 0.5 & 0.5 & 0.1 & 0.5 & 0.1 & 0.4 & 0.1 & 0.3 \\
\textbf{DFAUST} & 5k & 2k & 5e-3 & 1e-4 & 15 & 0.5 & 0.5 & 0.25 & 0.5 & 0.1 & 0.4 & 0.1 & 0.3 \\
\textbf{SMAL} & 4k & 2k & 5e-3 & 1e-4 & 10 & 0.5 & 0.5 & 0.25 & 0.5 & 0.1 & 0.4 & 0.1 & 0.3 \\
\bottomrule
\end{NiceTabular}
\end{minipage}%
\begin{minipage}{.25\textwidth}
    \begin{NiceTabular}{l ccc}[colortbl-like]
    \CodeBefore
      \columncolor{lightheadergrey}{3}
        \rowcolor{darkblue}{1}
    \Body
    \toprule
    Opt & $\boldsymbol{\lambda_1}$ & $\boldsymbol{\lambda_2}$ & $\boldsymbol{\lambda_3}$ \\
    \midrule
    \textbf{Main} & 2e2 & 1e1 & 5e3 \\
    \textbf{FT} & 1e3 & 1e2 & 5e3 \\
    \bottomrule
    \end{NiceTabular}
\end{minipage}
\caption{\textbf{Left:} Hyper-parameters \& Varifold settings used for training on each dataset. \textbf{Right:} Loss function weightings for the two stages of the optimisation; main \& fine tuning, across all datasets.}
\label{tab:combined_opt_parameters}
\end{table*}

\section{Quaternion Interpolation Derivation}

Section \ref{sec:skeleton_transformation} discusses how the locations of samples modelling bones are interpolated under quaternion rotations. In this section, we provide additional mathematical background on the derivation of these formulae.

\looseness=-1 Given a point \( \mathbf{p}_{0}  \) and a quaternion representing rotation \( \mathbf{q} \), the position of the point after the rotation has been applied is:
\begin{equation}
p_{1} = \mathbf{q} \cdot \mathbf{p}_{0} \cdot \mathbf{q}^{*}
\end{equation}
If we assume that over $t \in [0, 1]$ we apply a rotation of $\mathbf{q}$ and translation of $\mathbf{T}$, then the path traced out will be 
\begin{equation}
\mathbf{p} ( t ) = t \mathbf{T} + \mathbf{q}(t) \cdot \mathbf{p}_{0} \cdot \mathbf{q}^{*} (t)
\end{equation}
Now we want to find the velocity field $\mathbf{f_{q}}( \mathbf{p}, t)$ that will trace out the same trajectory for any initial $\mathbf{p}_0$. Thus we have 
\begin{equation}
\mathbf{p}_T = \mathbf{p}_0 + \int_{0}^{t} \mathbf{f}( \mathbf{p}, t) \,dt
\end{equation}
Therefore,
\begin{equation}
\int_{0}^{t} \mathbf{f}( \mathbf{p}, t) \,dt = \mathbf{p}(t) - \mathbf{p}_0 = t \mathbf{T} +  \mathbf{q}(t) \cdot \mathbf{p}_{0} \cdot \mathbf{q}^{*} (t)
\end{equation}
And hence,
\begin{equation}
\frac{\partial}{\partial t} \Rightarrow \mathbf{f}( \mathbf{p}, t) = \mathbf{T} + \frac{\partial}{\partial t}  \Bigl[ \mathbf{q}(t) \cdot \mathbf{p}_{0} \cdot \mathbf{q}^{*} (t)  \Bigr]
\end{equation}
By product rule,
\begin{equation}
\frac{\partial}{\partial t}  \Bigl[ \mathbf{q}(t) \cdot \mathbf{p}_{0} \cdot \mathbf{q}^{*} (t)  \Bigr] = \frac{\partial \mathbf{q} \left ( t \right)}{\partial t} \cdot \mathbf{p}_0 \cdot \mathbf{q}^{*} (t) + \mathbf{q}(t) \cdot \mathbf{p}_{0} \cdot \frac{\partial  \mathbf{q}^{*} (t)}{\partial t}
\end{equation}
Now, interpolating between between a unit identity quaternion and some new quaternion $\mathbf{q}$,
\begin{equation}
\mathbf{q} (t) = \text{SLERP} ( \mathbf{1}, \mathbf{q}_{0}, t )  = \left( \cos \frac{ \alpha t}{2}, \vv{\mathbf{n}} \sin \frac{ \alpha t}{2} \right) = \mathbf{q}_{0}^{t} \ , 
\end{equation}
where $0 \leq t \leq T$.

For a unit quaternion,
\begin{equation}
\mathbf{q}^{t} = \exp( \ln ( \mathbf{q} ) * t ) \quad 
\end{equation}
The derivative of the function $\mathbf{q}$, where $\mathbf{q}$ is a constant unit quaternion is,
\begin{equation}
\frac{\partial}{\partial t} \mathbf{q}^{t} = \ln ( \mathbf{q} ) \cdot \mathbf{q}^{t} = \ln ( \mathbf{q} ) \cdot \exp( \ln ( \mathbf{q} ) * t )
\end{equation}
\noindent, where the quaternion forms of $\exp$, $\log$ and to a power are,
\begin{align}
\exp(\mathbf{q}) &= \exp(s) \left(\begin{array}{c} \cos(|\mathbf{v}|) \\ \frac{\mathbf{v}}{|\mathbf{v}|} \sin(|\mathbf{v}|) \end{array}\right) \ .\\
\ln(\mathbf{q}) &= \left(\begin{array}{c} \ln(|\mathbf{q}|) \\ \frac{\mathbf{v}}{|\mathbf{v}|} \arccos\left(\frac{s}{|\mathbf{q}|}\right) \end{array}\right) \ .\\
\mathbf{q}^\mathbf{p} &= \exp(\ln(\mathbf{q}) \cdot \mathbf{p}) \ .
\end{align}
Note that if $p$ is in fact scalar, then the power is
\begin{equation}
\mathbf{q}^p = \exp(\ln(\mathbf{q}) * p) \ .
\end{equation}

\section{Sparse Nystr{\"o}m Varifold Approximation}

This section details the Sparse Nystr{\"o}m Approximation algorithm introduced by Paul \etal \cite{paul2024sparse}. 
We recall \cref{sec:varifold_matching} where we describe the varifold matching loss
\begin{equation}
\begin{aligned}
    \lossVarifold := \;&d(\mathbfcal{X}^{(T)}, \mathbfcal{Y}) \\
    = \;&\hilbert{\mu_{\mathbfcal{X}^{(T)} - \mu_{\mathbfcal{Y}}}}{\mu_{\mathbfcal{X}^{(T)}} - \mu_{\mathbfcal{Y}}} \\
    = \;&\hilbert{ \mu_{\mathbfcal{X}^{(T)}} }{ \mu_{\mathbfcal{X}^{(T)}} } 
    - 2 \hilbert{ \mu_{\mathbfcal{X}^{(T)}} }{ \mu_{\mathbfcal{Y}} } \\
     &\;+ \hilbert{ \mu_{\mathbfcal{Y}} }{ \mu_{\mathbfcal{Y}} } \ . \label{eqn:varifold_metric_details}
\end{aligned}
\end{equation}
Again, $\mathbfcal{X}^{(T)}$ denotes the set of vertices, normals and differential surface areas that are obtained at time $t=T$ from pushing $\mathbfcal{X}$ through the $\mathsf{ODE_{Solve}}$ in \cref{sec:diffeomorphic_flow}.
In practice, we do not need to calculate $\hilbert{ \mu_{\mathbfcal{Y}} }{ \mu_{\mathbfcal{Y}} }$ as it is a constant due to the target $\mathbfcal{Y}$ remaining unchanged.

For the first two terms in \cref{eqn:varifold_metric_details}, we use a discrete approximation of the inner product integrals in \cref{eqn:varifold_inner_product}
\vspace{-5pt}
\begin{equation}
    \langle \mu_{\mathbfcal{X}}, \mu_{\mathbfcal{Y}} \rangle \approx
    \sum_{i=1}^{I_\mathbfcal{X}} \sum_{i'=1}^{I_\mathbfcal{Y}} \kappa_{\xx}(\xx_i, \yy_{i'}) \, \kappa_{\nn}(\sn_{\mathbfcal{X}_i}, \sn_{\mathbfcal{Y}_{i'}}) s_{\mathbfcal{X}_i} \, s_{\mathbfcal{Y}_{i'}} \ , \label{eqn:supp_discrete_varifold_product}
\end{equation}
where the summation has the computational complexity $\mathcal{O}(I_\mathbfcal{X}\,I_\mathbfcal{Y})$; this cost becomes very expensive when fitting to a very high resolution target (\eg fitting a template to a raw point cloud scan) where the number of vertices on the target is far larger than the template, $I_\mathbfcal{Y} \gg I_\mathbfcal{X}$ and we seek to reduce this cost.

\paragraph{Sparse Approximation}
The algorithm of Paul~\etal~\cite{paul2024sparse} creates sparse approximations of Varifold representations significantly smaller than the input data while maintaining high accuracy. 
It employs a Ridge Leverage Score (RLS) approach to assess a data point's importance, which is used to create a Nystr{\"o}m approximation for the in the Reproducing Kernel Hilbert Space (RKHS), offering a computationally efficient and accurate approach to shape compression. For a comprehensive understanding of the theoretical foundations, including detailed mathematical proofs, readers are referred to the original paper.

Application to the target, results in a compressed approximation $\mathbfcal{Y}^{c} = \{ V_\mathbfcal{S_{\mathbfcal{Y}^{c}}}, N_\mathbfcal{S_{\mathbfcal{Y}^{c}}}, \boldsymbol{\beta} \} $ containing a subset of $I_\mathbfcal{Y}^{c}$ of the original vertices and normals with a corresponding set of weights $\boldsymbol{\beta}$.
The Varifold matching representation from \cref{eqn:supp_discrete_varifold_product} to a compressed target becomes 
\begin{equation}
    \langle \mu_{\mathbfcal{X}}, \mu_{\mathbfcal{Y}^{c}} \rangle \approx
    \sum_{i=1}^{I_\mathbfcal{X}} \sum_{i'=1}^{I_{\mathbfcal{Y}^{c}}} \kappa_{\xx}(\xx_i, \yy^{c}_{i'}) \kappa_{\nn}(\sn_{\mathbfcal{X}_i}, \sn_{\mathbfcal{Y}^{c}_{i'}}) s_{\mathbfcal{X}_i} \, \mathbfcal{\beta}_{\mathbfcal{Y}^{c}_{j}} \ , \label{eqn:compressed_discrete_varifold_product}
\end{equation}
where $I_\mathbfcal{Y}^{c} \ll I_\mathbfcal{Y}$ dramatically reducing the computational cost of calculating the Varifold loss required.

\begin{algorithm}
\caption{Varifold Compression Algorithm}
\SetAlgoLined
\KwIn{\\ $\mathbfcal{Y}$: Uncompressed Data - $ \left\{ V_\mathbfcal{Y}, N_\mathbfcal{Y}, dS_\mathbfcal{Y} \right\}$ \\ $n$: Number of Samples in $\mathbfcal{Y}$ ($= I_\mathbfcal{Y}$) \\ $m$: Desired Compressed Size ($= I_{\mathbfcal{Y}^{c}}$) \\ $\lambda$: Regularisation parameter \\ $\kappa_{\xx}(\cdot, \cdot)$: Positional Kernel \\ $\kappa_{\nn}(\cdot, \cdot)$: Normal Kernel} 
\KwOut{\\
$\mathbfcal{Y}^{c}$ : Compressed Representation - $\left\{ V_{\mathbfcal{Y}^{c}} , N_{\mathbfcal{Y}^{c}}, \boldsymbol{\beta} \right\} $
}
\BlankLine
\setlength{\fboxsep}{1pt}

\colorbox{algosection}{
  \parbox{\dimexpr0.90\linewidth-2\fboxsep\relax}{
    \textcolor{sectioncolor}{$\blacktriangleright$ \textbf{Compute leverage scores:}}\\
    $b_{s} \gets \lfloor\sqrt{n}\rfloor$\; 
    $b \gets \lfloor n/s\rfloor$\; 
    Split $\mathbfcal{Y}$ into $b$ random batches $\left\{ \mathbfcal{Y}_1, \ldots, \mathbfcal{Y}_b \right\}$ of size $b_{s}$\;
    \For{$j = 1$ \KwTo $b$}{
        \For{$i = 1$ \KwTo $b_{s}$}{
            $\boldsymbol{\Lambda}_i \gets \mathbf{K}_{i, \mathbfcal{Y}_{j}} {( \mathbf{K}_{i, \mathbfcal{Y}_{j}} + \lambda \mathbf{I})}^{-1} $\;
            \colorbox{algosectiondarker}{
                \parbox{\dimexpr0.82\linewidth-2\fboxsep\relax}{
                    \small $\blacktriangleright \mathbf{K}_{i,\mathbfcal{Y}_{j}} = \sum_{i' \in \mathbfcal{Y}_{j}} \! 
                    \kappa_{\xx}(\yy_i, \yy_{i'}) \kappa_{\nn}(\sn_{\mathbfcal{Y}_{i}}, \sn_{\mathbfcal{Y}_{i'}})$
                }
            }
        }
    }
  }
}

\BlankLine
\colorbox{algosection}{
  \parbox{\dimexpr0.9\linewidth-2\fboxsep\relax}{
    \textcolor{sectioncolor}{$\blacktriangleright$ \textbf{Draw weighted samples:}}\\
    Define sampling distribution:\\
    Let $X_i \sim p(W)$ where $p(X_i = s_j) = \frac{\boldsymbol{\Lambda}_j}{\sum_{k=1}^n \boldsymbol{\Lambda}_k}$\;
    $\mathbfcal{C} \gets \left\{ \ \right\}$\;
    \For{$i = 1$ \KwTo $m$}{
        $\left\{\xx_{s}, \sn_{\mathbfcal{Y}_{s}} \right\} \gets$ Draw a sample from $\mathbfcal{Y}$ acc. to $p(W)$\;
        Add $\left\{\xx_{s}, \sn_{\mathbfcal{Y}_{s}} \right\}$ to $\mathbfcal{C}$\;
    }
  }
}

\BlankLine
\colorbox{algosection}{
  \parbox{\dimexpr0.9\linewidth-2\fboxsep\relax}{
    \textcolor{sectioncolor}{$\blacktriangleright$ \textbf{Calculate sample weights:}}\\
    $\boldsymbol{\beta} \gets {\mathbf{K}^{-1}_{\mathbfcal{C}, \mathbfcal{C}}} \mathbf{Y} $\;
    \colorbox{algosectiondarker}{
                \parbox{\dimexpr0.99\linewidth-2\fboxsep\relax}{
                    \small $\blacktriangleright \mathbf{K}_{\mathbfcal{C}, \mathbfcal{C}} = \sum_{i=1}^{I_\mathbfcal{C}} \sum_{i'=1}^{I_\mathbfcal{C}} \kappa_{\xx}(\mathbf{c}_i, \mathbf{c}_{i'}) \kappa_{\nn}(\sn_{\mathbfcal{C}_i}, \sn_{\mathbfcal{C}_{i'}})$ \\
                    \small $\blacktriangleright \mathbf{Y} =
    \sum_{i=1}^{I_\mathbfcal{C}} \sum_{i'=1}^{I_\mathbfcal{Y}} \kappa_{\xx}(\mathbf{c}_i, \xx_{i'}) \kappa_{\nn}(\sn_{\mathbfcal{C}_i}, \sn_{\mathbfcal{Y}_{i'}}) s_{\mathbfcal{Y}_i} \, s_{\mathbfcal{Y}_{i'}}$
            }
        }
  }
}
\BlankLine
\KwRet{ $\left\{ V_{\mathbfcal{Y}^{c}}, N_{\mathbfcal{Y}^{c}}, \boldsymbol{\beta} \right\}$ }
\label{al:varifold_compression}
\end{algorithm}

\paragraph{Compression Process}
The compression process, as outlined in \Cref{al:varifold_compression}, consists of three main steps: 

\begin{enumerate}
\item The RLS values for each input element are generated efficiently using a sampling method that avoids calculating the full all-pairwise matrices. \item Control points are then selected using a weighted sampling approach, with the RLS score determining the probability of selection.
\item Updated weights are calculated for each selected control point. 
\end{enumerate}

Several parameters are required as input for the compression process. Firstly, the desired compression size $m < I_\mathbfcal{Y}$ determines the final number of control points. Additionally, the length scales of the kernels, $\kappa_{\xx}$ and $\kappa_{\nn}$, need to be set (they are the same as in the matching algorithm $\ell_\xx$ and $\ell_\nn$). Finally, an approximation parameter $\lambda$ is required; we used a default value of 1 for all our experiments.

\section{Additional results}

This section presents additional results that were omitted from the main results section due to space limitations.

\subsection{More Quantitative Results}
\label{sec:additional_quantitative_results}

\begin{figure*}[p!]
\centering
\begin{minipage}[t][0.85\linewidth][b]{0.02\linewidth}
\vfill\begin{sideways} \textbf{Ours} \end{sideways}\vfill%
\vfill\begin{sideways} \textbf{SMS}~\protect\cite{cao2024spectral} \end{sideways}\vfill%
\vfill\begin{sideways} \textbf{ESA}~\protect\cite{hartman2023elastic} \end{sideways}\vfill\phantom{a}%
\end{minipage}%
\begin{minipage}[t][0.85\linewidth][b]{0.98\linewidth}
\begin{tabular}{>{\centering\arraybackslash}p{0.305\linewidth}>{\centering\arraybackslash}p{0.305\linewidth}>{\centering\arraybackslash}p{0.305\linewidth}}
\textbf{Geodesic Distance} & \textbf{Chamfer Distance} & \textbf{Quasi-Conformal Distortion}
\end{tabular}\\
\includegraphics[width=\linewidth]{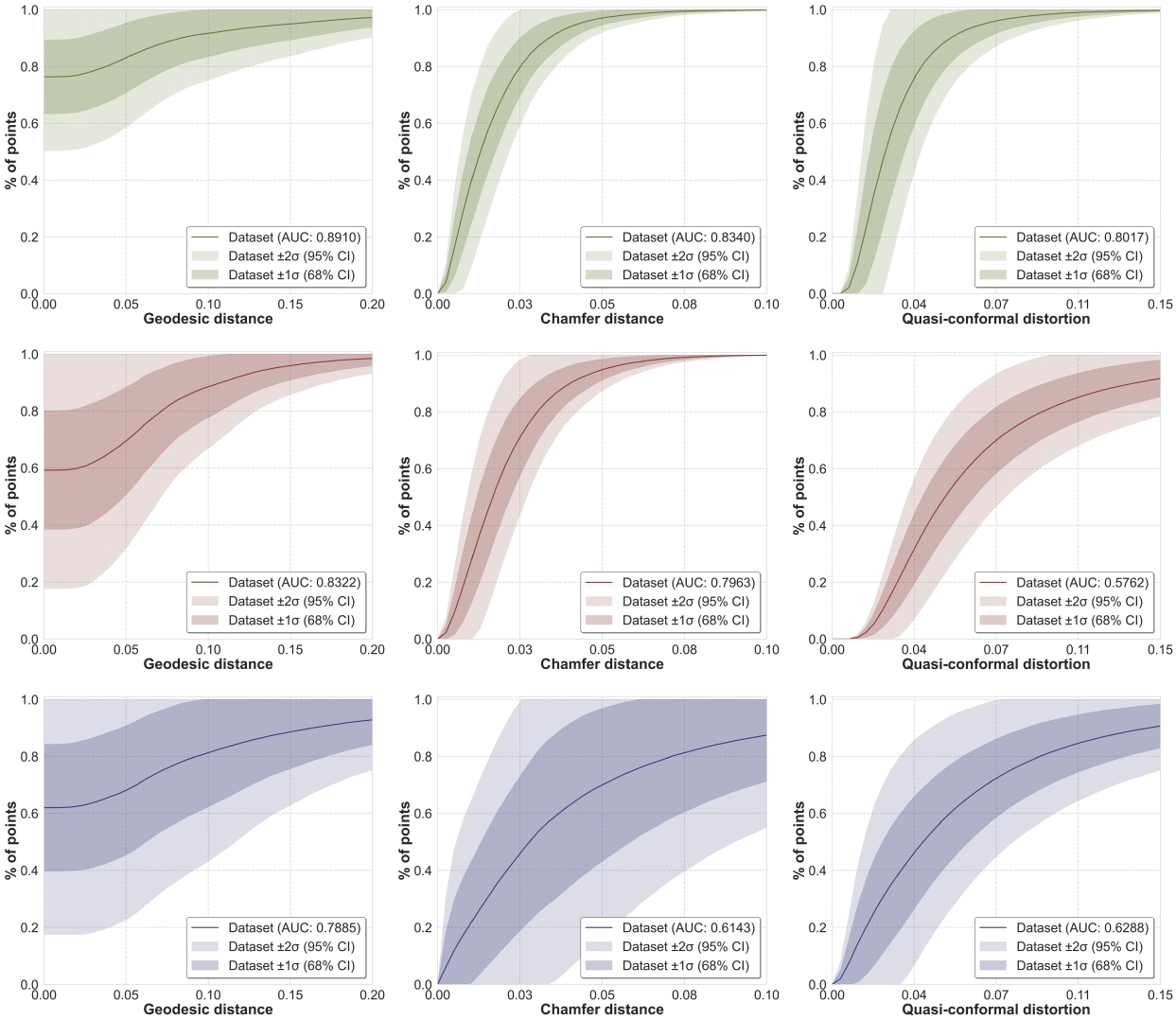}
\end{minipage}%
    \caption{\textbf{Interpolation results for MANO: Ours vs SMS~\protect\cite{cao2024spectral} \& ESA ~\protect\cite{hartman2023elastic}.} Mean and confidence intervals for the three metrics are shown; top row has our results, middle SMS \& bottom row ESA.}
    \label{fig:mano_combined_results}
\end{figure*}

\begin{figure*}[p!]
\centering
\begin{minipage}[t][0.85\linewidth][b]{0.02\linewidth}
\vfill\begin{sideways} \textbf{Ours} \end{sideways}\vfill%
\vfill\begin{sideways} \textbf{SMS}~\protect\cite{cao2024spectral} \end{sideways}\vfill%
\vfill\begin{sideways} \textbf{ESA}~\protect\cite{hartman2023elastic} \end{sideways}\vfill\phantom{a}%
\end{minipage}%
\begin{minipage}[t][0.85\linewidth][b]{0.98\linewidth}
\begin{tabular}{>{\centering\arraybackslash}p{0.305\linewidth}>{\centering\arraybackslash}p{0.305\linewidth}>{\centering\arraybackslash}p{0.305\linewidth}}
\textbf{Geodesic Distance} & \textbf{Chamfer Distance} & \textbf{Quasi-Conformal Distortion}
\end{tabular}\\
\includegraphics[width=\linewidth]{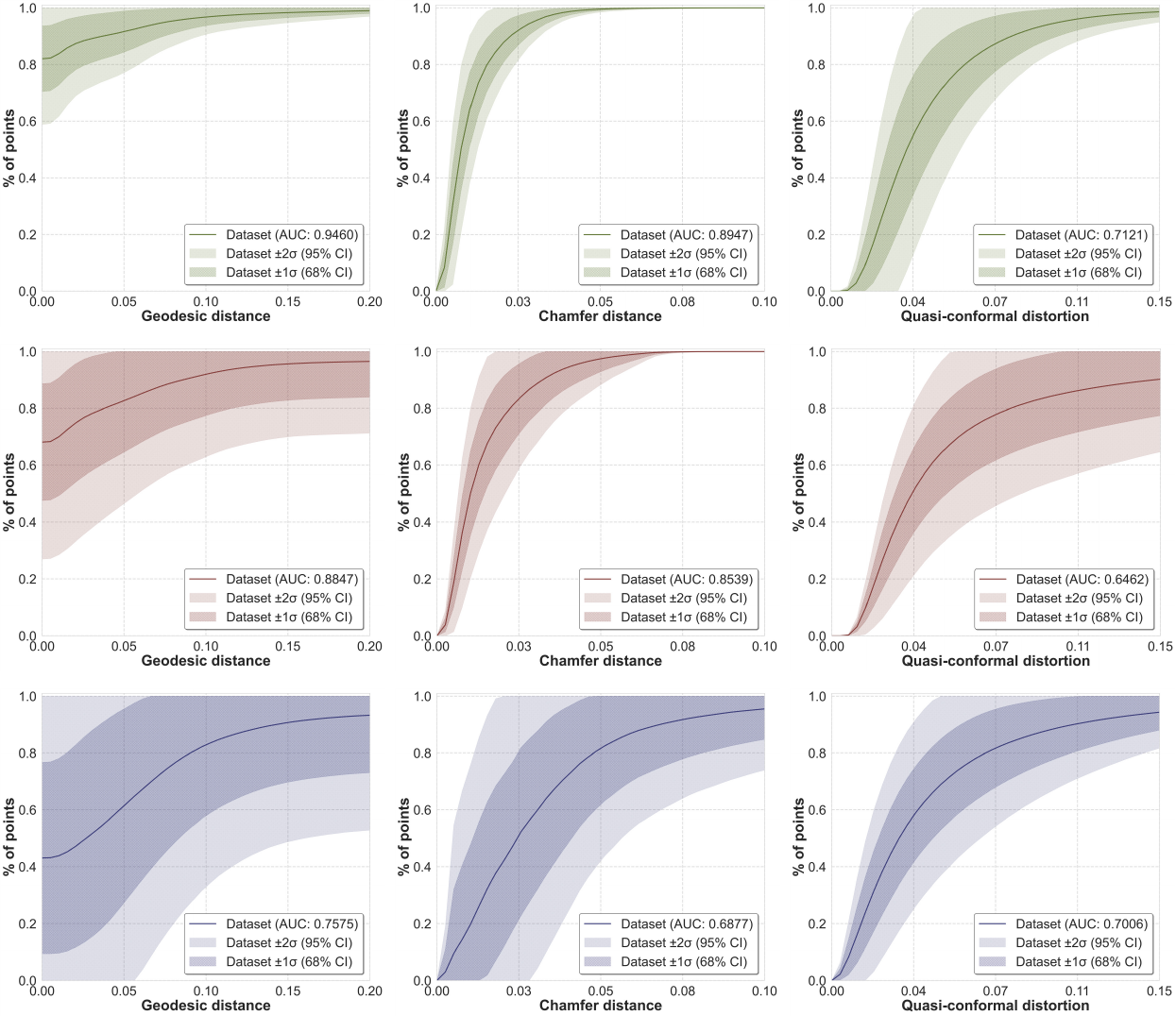}
\end{minipage}%
    \caption{\textbf{Interpolation results for DFAUST: Ours vs SMS~\protect\cite{cao2024spectral} \& ESA ~\protect\cite{hartman2023elastic}.} Mean and confidence intervals for the three metrics are shown; top row has our results, middle SMS \& bottom row ESA.}
    \label{fig:dfaust_combined_results}
\end{figure*}

\begin{figure*}[p!]
\centering
\begin{minipage}[t][0.85\linewidth][b]{0.02\linewidth}
\vfill\begin{sideways} \textbf{Ours} \end{sideways}\vfill%
\vfill\begin{sideways} \textbf{SMS}~\protect\cite{cao2024spectral} \end{sideways}\vfill%
\vfill\begin{sideways} \textbf{ESA}~\protect\cite{hartman2023elastic} \end{sideways}\vfill\phantom{a}%
\end{minipage}%
\begin{minipage}[t][0.85\linewidth][b]{0.98\linewidth}
\begin{tabular}{>{\centering\arraybackslash}p{0.305\linewidth}>{\centering\arraybackslash}p{0.305\linewidth}>{\centering\arraybackslash}p{0.305\linewidth}}
\textbf{Geodesic Distance} & \textbf{Chamfer Distance} & \textbf{Quasi-Conformal Distortion}
\end{tabular}\\
\includegraphics[width=\linewidth]{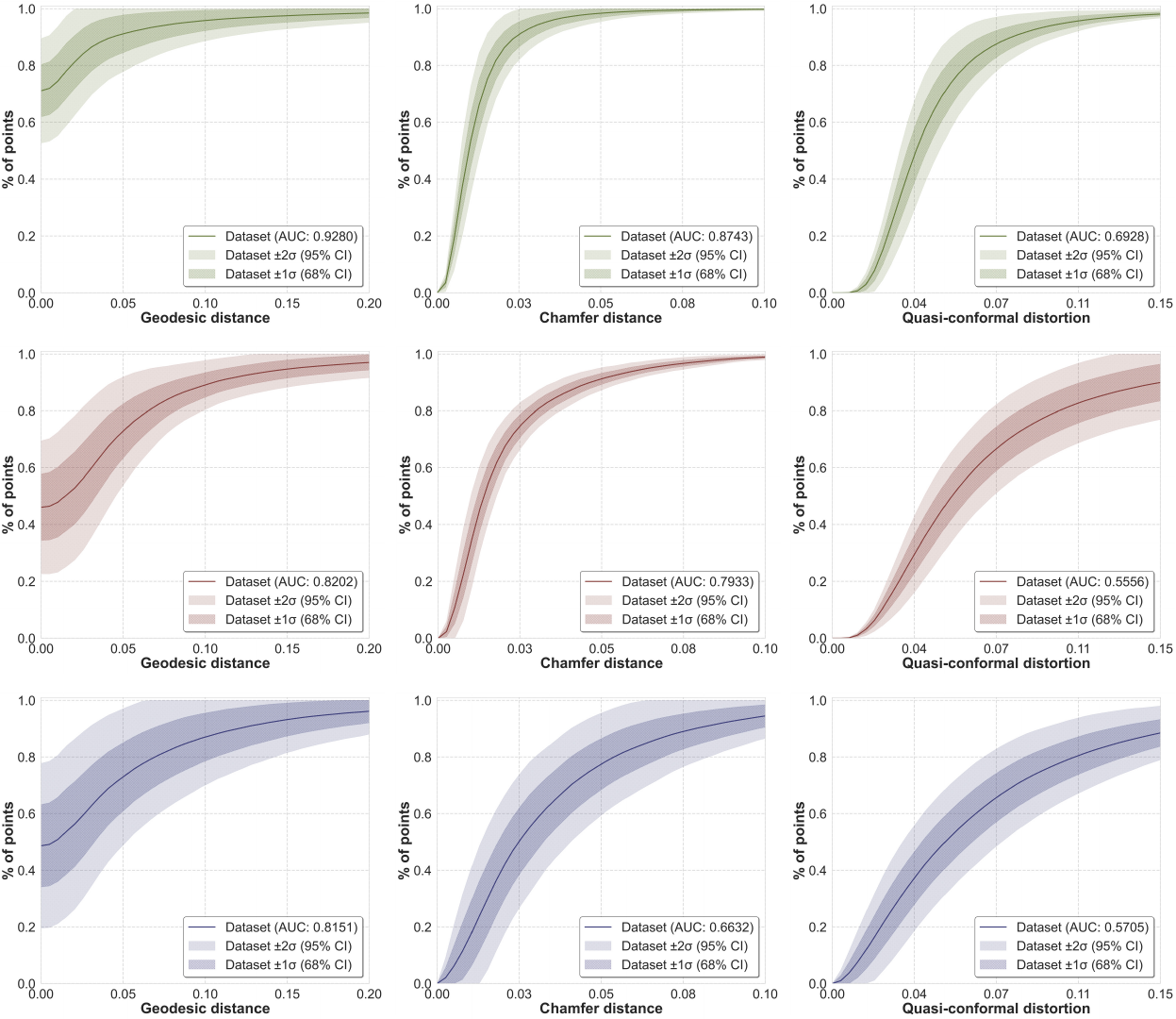}
\end{minipage}%
    \caption{\textbf{Interpolation results for SMAL: Ours vs SMS~\protect\cite{cao2024spectral} \& ESA ~\protect\cite{hartman2023elastic}.} Mean and confidence intervals for the three metrics are shown; top row has our results, middle SMS \& bottom row ESA.}
    \label{fig:smal_combined_results}
\end{figure*}

In \Cref{fig:mano_combined_results,fig:dfaust_combined_results,fig:smal_combined_results} the interpolation results for our method against the state-of-the-art SMS \cite{cao2024spectral} and ESA \cite{hartman2023elastic} methods for the three datasets; MANO, DFAUST and SMAL, showing the mean and confidence intervals for all metrics. 

Our method demonstrates improvement across all metrics for each dataset, showing both higher mean values and reduced variance in results. 

\begin{figure*}[p!]
\centering
\begin{minipage}[t][0.85\linewidth][b]{0.02\linewidth}
\vfill\begin{sideways} \textbf{Ours} \end{sideways}\vfill%
\vfill\begin{sideways} \textbf{SMS}~\protect\cite{cao2024spectral} \end{sideways}\vfill%
\vfill\begin{sideways} \textbf{ESA}~\protect\cite{hartman2023elastic} \end{sideways}\vfill\phantom{a}%
\end{minipage}%
\begin{minipage}[t][0.85\linewidth][b]{0.98\linewidth}
\begin{tabular}{>{\centering\arraybackslash}p{0.305\linewidth}>{\centering\arraybackslash}p{0.305\linewidth}>{\centering\arraybackslash}p{0.305\linewidth}}
\textbf{Geodesic Distance} & \textbf{Chamfer Distance} & \textbf{Quasi-Conformal Distortion}
\end{tabular}\\
\includegraphics[width=\linewidth]{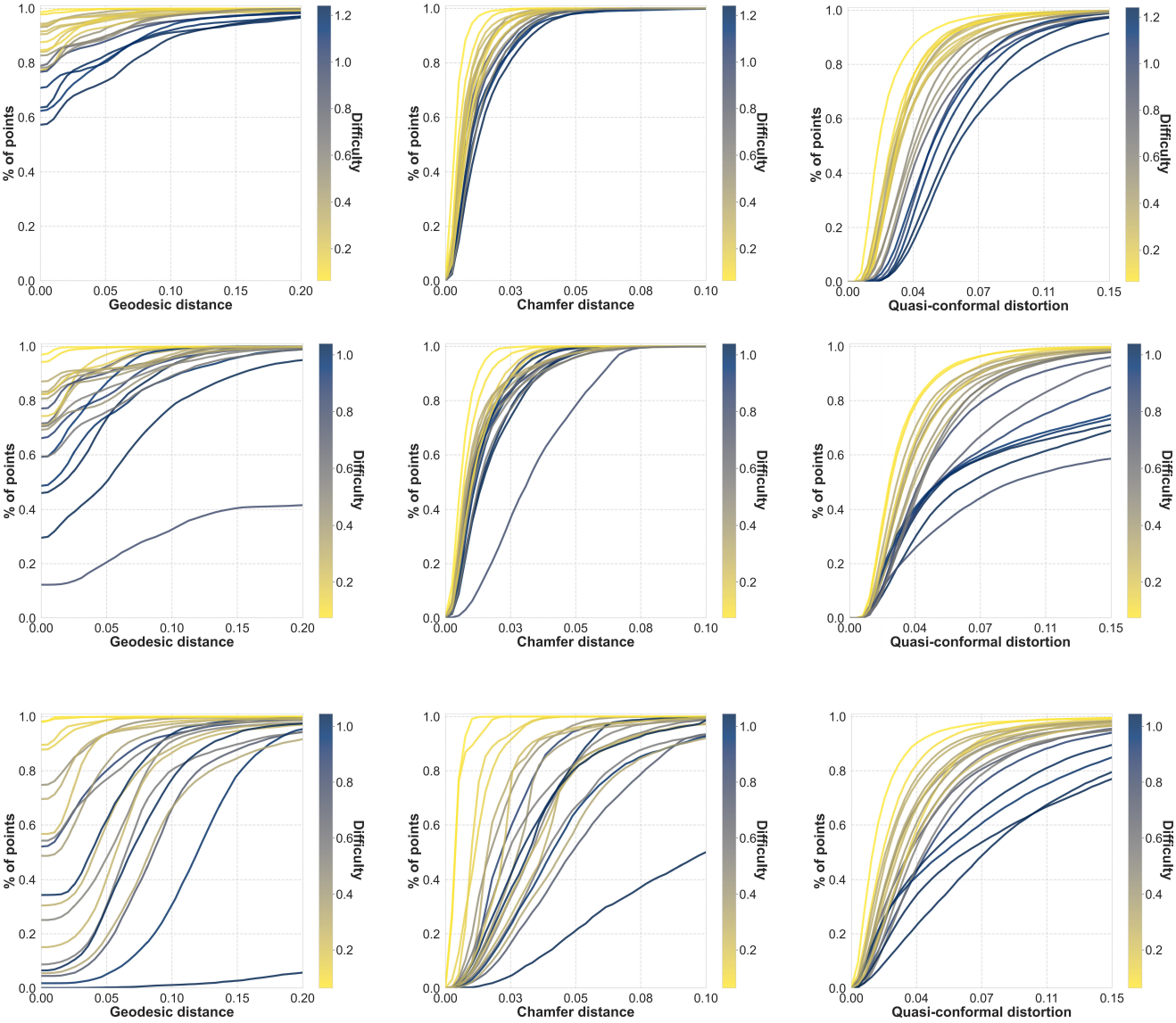}
\end{minipage}%
    \caption{\textbf{Interpolation results for DFaust: Ours vs SMS~\protect\cite{cao2024spectral} \& ESA~\protect\cite{hartman2023elastic}.} Plotting individual interpolations in which the difficulty of the problem is colour-coded; top row has our results, middle SMS \& bottom row ESA. The difficulty rating is determined by the average vertex displacement for each interpolation task (from the ground-truth) normalised to one.}
    \label{fig:difficulty_plots}
\end{figure*}

The reduction in variance of our method can be attributed primarily to our method's superior performance on more challenging problems. This is illustrated in \cref{fig:difficulty_plots}, where we plot individual curves for interpolations where we colour each line with a ``difficulty rating'' calculated based on the average vertex displacement (normalised to one). 
When dealing with shapes that undergo a larger degree of deformation, competing methods show a significant drop in the quality of their results; in contrast, our approach maintains its effectiveness, leading to more consistent performance across varying levels of problem difficulty. All approaches show a roughly monotone increase in performance as the difficulty rating decreases.

\subsection{Further Qualitative Results}

In this section we provide additional qualitative results, highlighting some of the advantages of our method which may not be accounted for by performance metrics alone. 

\subsection{Non-Intersection of Surfaces}

\begin{figure*}[t!]
    \centering
   \includegraphics[width=\linewidth]{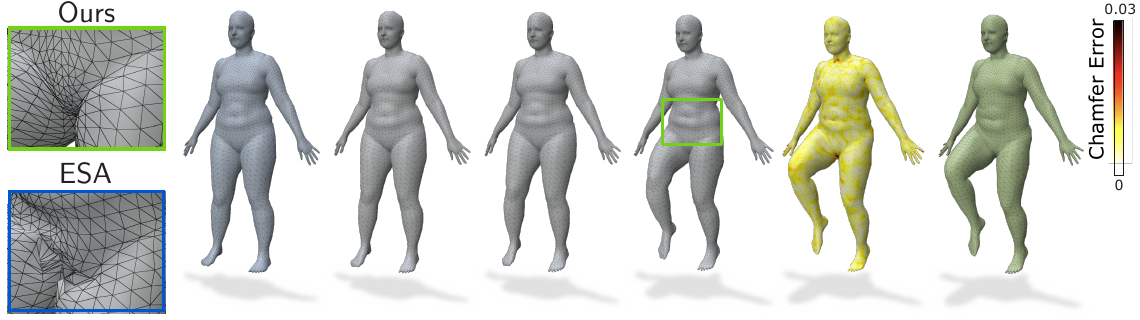}
       \vspace{-20pt}
    \caption{\textbf{Non-Intersection of Surfaces:} We show interpolants from our method from the source on the left to target on the right (with final chamfer error on the second to right). As the leg is raised the interpolation from ESA fails to maintain surface integrity where the top of the leg meets the stomach resulting in unrealistic mesh distortions and overlapping (see closeup on bottom-left). Our diffeomorphic approach preserves topology by construction and guarantees non-intersection of the mesh (see closeup on top-left).}
    \vspace{-10pt}
    \label{fig:self_intersection_example}
\end{figure*}

In \cref{fig:self_intersection_example}, we demonstrate how our method handles a leg lift scenario where the leg comes into contact with the stomach of the individual. Since we model deformation as a diffeomorphism, represented by a time-varying vector flow field, our approach guarantees non-intersection by construction. In contrast, the interpolation produced by the ESA method fails to maintain surface integrity, resulting in unrealistic overlapping and severe mesh distortions.

\pagebreak
\newpage

\pagebreak
\newpage
\FloatBarrier

\begin{figure*}[b!]
    \vspace{-10pt}
    \centering
   \includegraphics[width=\linewidth]{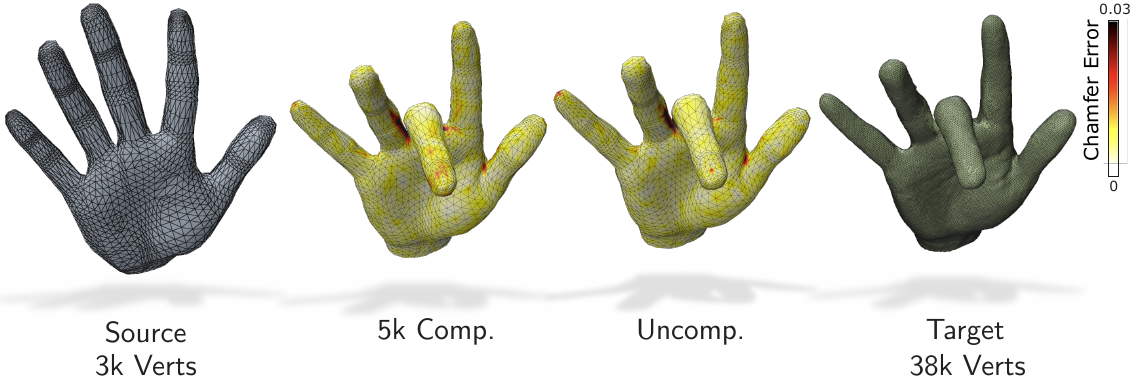}
       \vspace{-10pt}
    \caption{\textbf{Fitting Template to Raw Scan Data.} We use a high-resolution noisy scan as the target to illustrate use of our method for template fitting. Varifold compression improves the efficiency of our approach with negligible change in final quality to using the full (uncompressed) target. Our method is robust to the noisy and partial data found in the dense target scan.}
    \vspace{-10pt}
    \label{fig:mano_template_fitting}
\end{figure*}


\subsection{Fitting a Template to Topologically Noisy Data}

A common approach in statistical shape analysis is to first bring all the raw data into correspondence by fitting a template to each sample. This also has the advantage that it removes noise and partial surfaces. However, this is a tricky task that often requires manual input. 

We present an example demonstrating the potential for our method to automate this task. We select a neutral pose from the MANO dataset as our ``template'' and attempt to register this to a raw hand scan consisting of 38k vertices. Using the compression of Paul \etal \cite{paul2024sparse}, we form a 5k compressed representation as the target (in increasing computational efficiency).

\Cref{fig:mano_template_fitting} illustrates the result of applying our approach, resulting in a high-quality registration. Although the majority of the surface fit has a very low Chamfer error, an area of higher error can be observed on the ring finger. This is due to the noise in the raw scan, where an unnatural bulge is clearly visible on one of the fingers. The use of a volume-preserving constraint by construction allows our method to fit the template despite this noise in the raw data. 

Overall, as previously demonstrated there is no difference in the quality of the fit between using the full raw data and a Varifold compressed representation.

\subsection{Automatic Skeleton Transfer}

\begin{figure*}[t!]
\centering
\begin{minipage}[t]{0.79\textwidth}
    \centering
   \includegraphics[width=\linewidth]{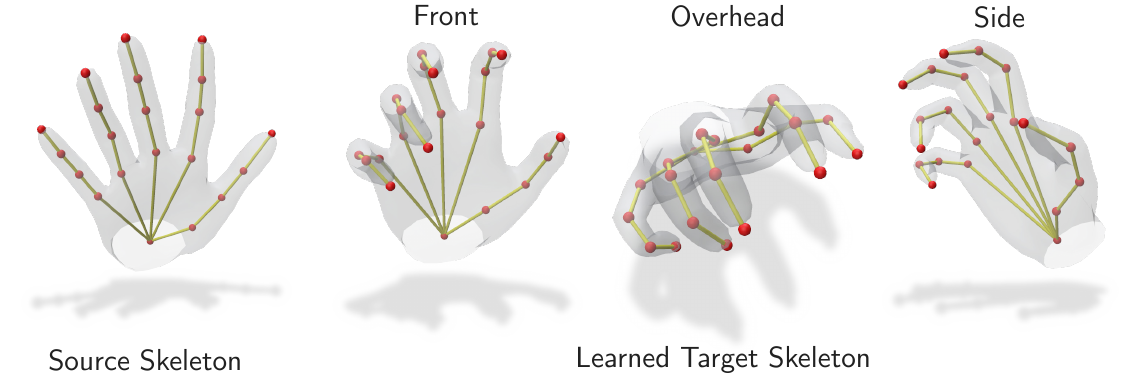}
\end{minipage}%
    \vspace{0.1em} 
    \centerline{\color{gray!30}\rule{\columnwidth}{0.5pt}}
\begin{minipage}[t]{0.79\textwidth}
    \centering
   \includegraphics[width=\linewidth]{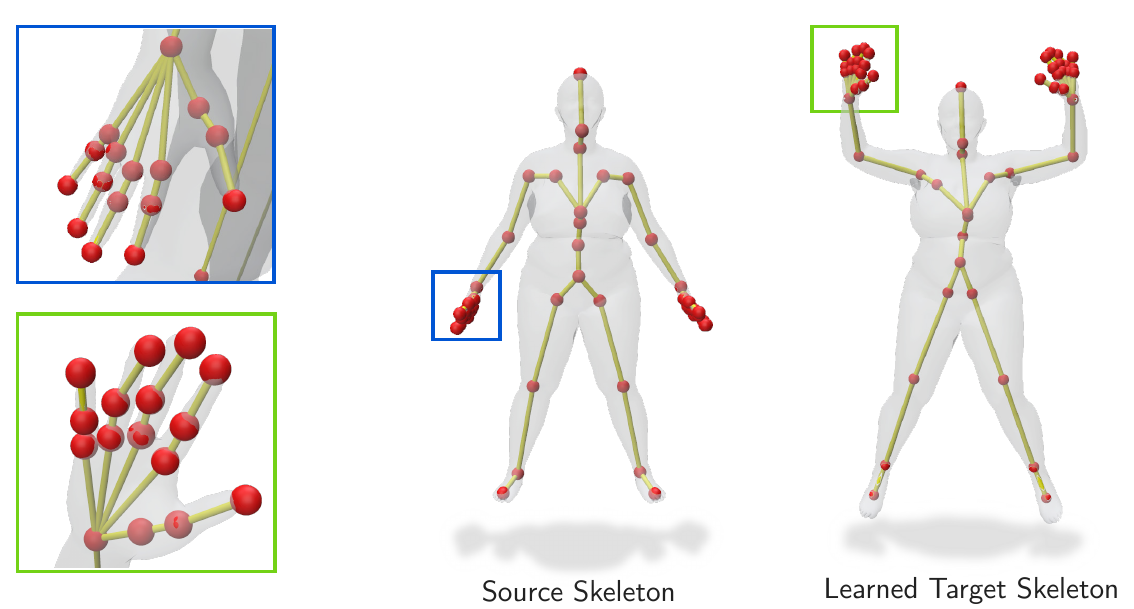}
\end{minipage}
    \vspace{0.5em}
    \centerline{\color{gray!30}\rule{\columnwidth}{0.5pt}}
\begin{minipage}[t]{0.79\textwidth}
    \centering
   \includegraphics[width=\linewidth]{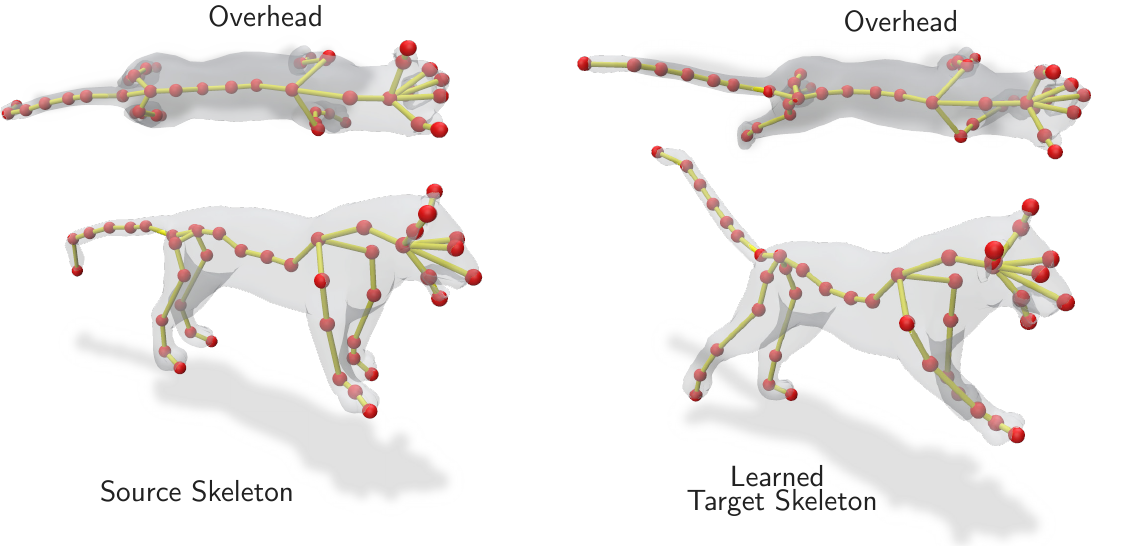}
\end{minipage}
    \caption{\textbf{Learned Target Skeleton Examples:} Skeletons resulting from interpolations between poses involving \textbf{Top:} MANO, \textbf{Middle:} DFAUST \& \textbf{Bottom:} SMAL datasets. The target skeleton is learned as a by-product of our method without any prior knowledge of the ground truth.}
    \label{fig:skeleton_transfer}
\end{figure*}

The animation community has spent significant effort trying to ease rigging procedures. This is necessitated because the increasing availability of 3d data makes manual rigging infeasible. However, object animations involve understanding elaborate geometry and dynamics, and such knowledge is hard to infuse even with modern data-driven techniques. Automatic rigging methods do not provide adequate control and cannot generalise in the presence of unseen artefacts. An alternative approach is to learn to transfer an existing rig to a target using a dataset of known target poses to train 
a neural network. 

\vfill 
\balance
\newpage
\FloatBarrier

\begin{figure*}[t!]
\centering
   \includegraphics[width=\linewidth]{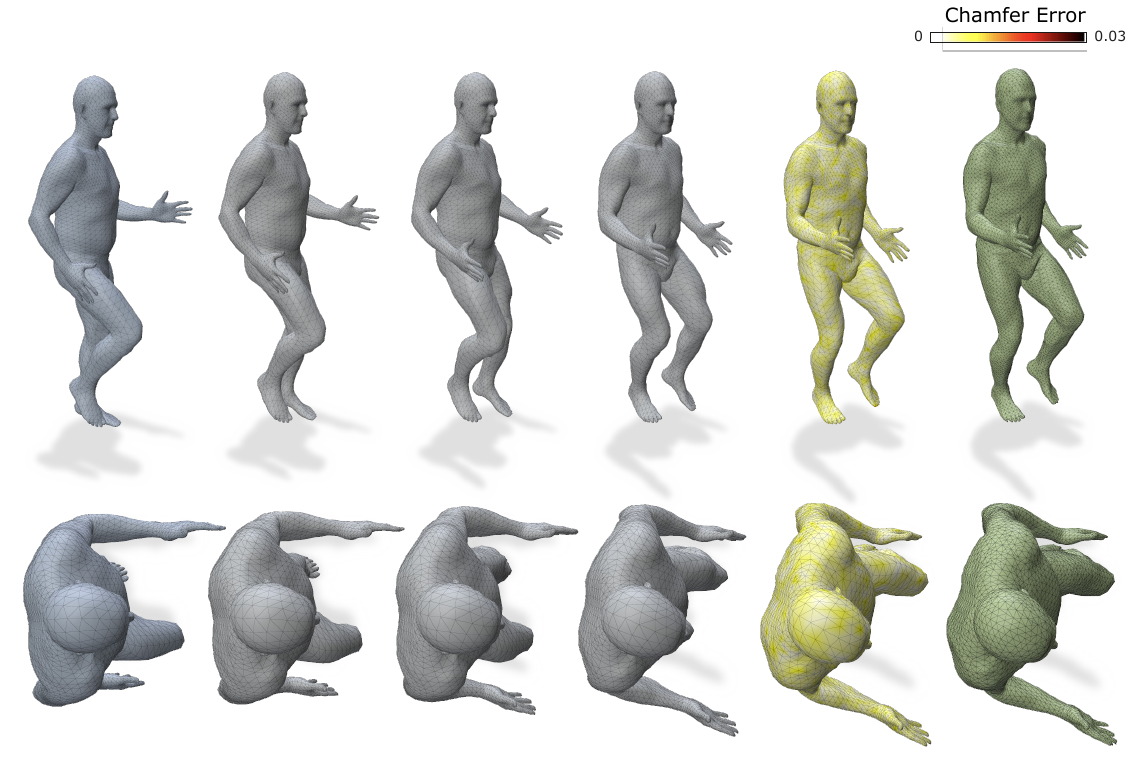}
       \vspace{-20pt}
       \caption{\textbf{Interpolation Between Frames Capturing A Man Running on the Spot From DFAUST Dataset:} We show interpolants from our method from the source on the left to target on the right (with final chamfer error on the second to right). }
    \label{fig:final_interpolations_dfaust}
\end{figure*}

\flushbottom

As part of our method the forward kinematics of the final skeletal pose (the global translation $\tilde\trans \in \mathbb{R}^3$ and quaternion joint angles $\tilde\quat_k \in \mathbb{Q}$ are optimised. As a result, we learn to transfer the source skeleton to the target as a by-product of our method. In \Cref{fig:skeleton_transfer} we provide examples of these transferred skeletons, which appear in plausible configurations, and notably were achieved without any prior knowledge of ground truth target configurations.

\subsection{Additional Interpolations Examples}

\begin{figure*}[t!]
    \centering
   \includegraphics[width=\linewidth]{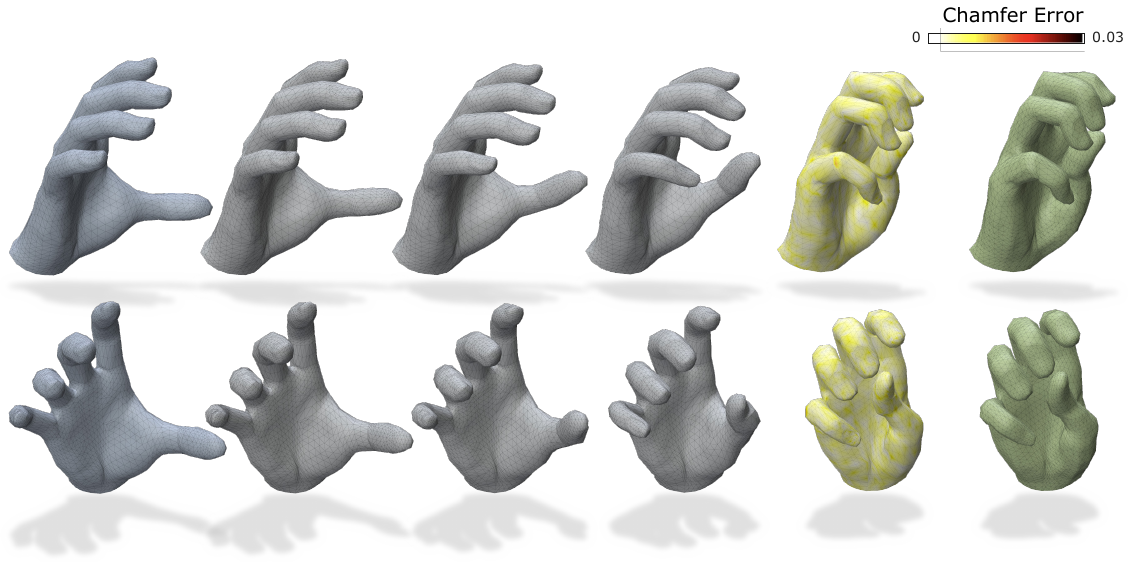}
   \vspace{-15pt}
       \caption{\textbf{Interpolation Between An Open \& Closed Hand Poses from the MANO Dataset:} We show interpolants from our method from the source on the left to target on the right (with final chamfer error on the second to right).}
       \label{fig:final_interpolations_mano}
           \vspace{0.5em}
    \centerline{\color{gray!30}\rule{\columnwidth}{0.5pt}}
\end{figure*}

\begin{figure*}[b!]
    \centering
    \centering
   \includegraphics[width=\linewidth]{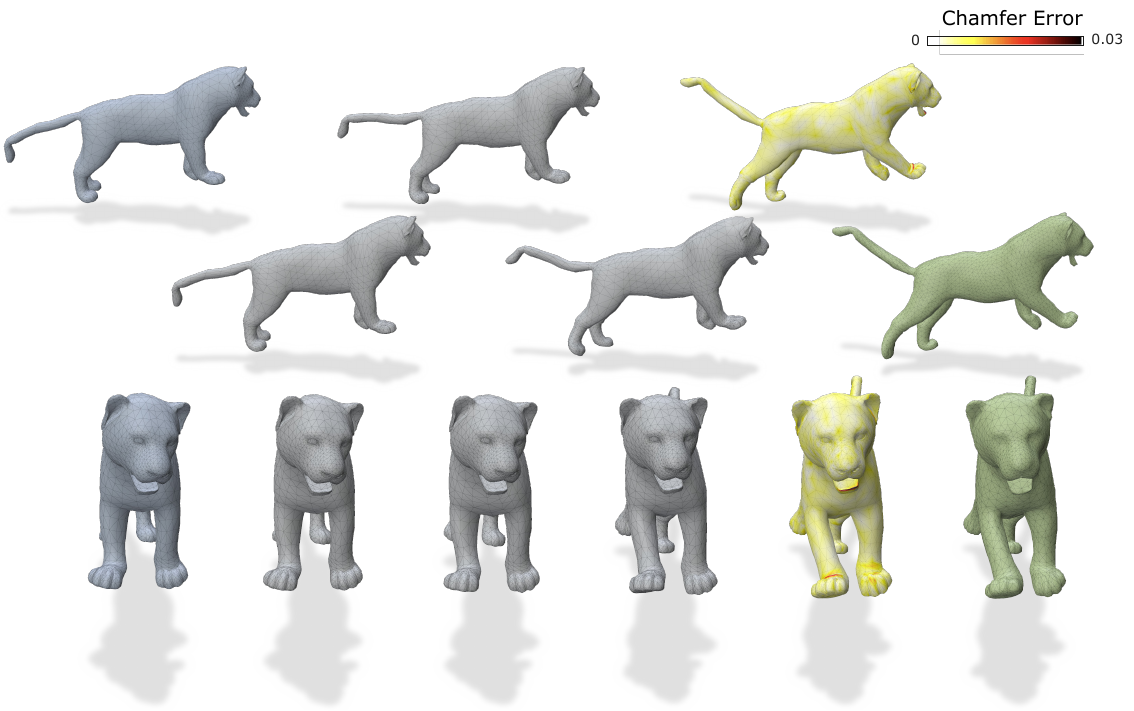}
   \vspace{-15pt}
    \caption{\textbf{Interpolation Between Two Running Poses from the SMAL Dataset:} We show interpolants from our method from the source on the left to target on the right (with final chamfer error on the second to right).}
       \label{fig:final_interpolations_smal}
\end{figure*}

To further illustrate our findings, in \Cref{fig:final_interpolations_dfaust,fig:final_interpolations_mano,fig:final_interpolations_smal} we present additional interpolation examples generated using our method as a comprehensive showcase 
of our technique's capabilities.

\newpage

\end{document}